\documentclass{article} 
\pdfoutput=1
\usepackage[preprint]{colm2025_conference}

\usepackage{amsmath,amsfonts,amssymb,bm}

\def\eqref#1{equation~\ref{#1}}

\def\1{\bm{1}}

\DeclareMathAlphabet{\mathsfit}{\encodingdefault}{\sfdefault}{m}{sl}
\SetMathAlphabet{\mathsfit}{bold}{\encodingdefault}{\sfdefault}{bx}{n}

\usepackage{microtype}
\usepackage{hyperref}
\usepackage{url}
\usepackage{booktabs}
\usepackage{lineno}

\usepackage{textcomp}
\usepackage{fontawesome}
\usepackage{multirow}
\usepackage[color=blue!50]{todonotes}
\usepackage{graphicx}
\usepackage{array}
\usepackage{listings}
\usepackage{caption}
\usepackage{subcaption}
\usepackage{pmboxdraw}
\usepackage{wrapfig}
\usepackage{ifthen}
\usepackage{enumitem}
\usepackage[breakable]{tcolorbox}
\usepackage{bm}

\definecolor{darkblue}{rgb}{0, 0, 0.5}
\hypersetup{colorlinks=true, citecolor=darkblue, linkcolor=darkblue, urlcolor=darkblue}
\definecolor{teal}{HTML}{085c64}

\newboolean{owa}
\setboolean{owa}{false} 
\newcommand{\rewrite}[1]{{\color{black}#1}}
\newcommand{\verbose}[1]{{\color{black}#1}}
\newcommand{\change}[1]{{\color{black}#1}}
\newcommand{\added}[1]{{\color{black}#1}}
\newcommand{\cready}[1]{{\color{black}#1}}

\newcommand{\github}{\href{https://github.com/google/feabench/tree/main}{\faGithub}}

\newcommand\aref[1]{Appendix \ref{#1}}
\newcommand\sref[1]{Section \ref{#1}}
\newcommand\fref[1]{Figure \ref{#1}}
\newcommand\tref[1]{Table \ref{#1}}
\newcommand{\regc}{\textsuperscript{\textregistered}}
\newcommand{\comsol}{COMSOL Multiphysics\regc }
\newcommand{\benchmain}{FEABench Gold}
\newcommand{\benchlarge}{FEABench Large}

\newcommand{\model}{\textbf{ModelSpecs} }
\newcommand{\plan}{\textbf{Plan} }

\definecolor{lm}{HTML}{5954a4}
\definecolor{corr}{HTML}{6aa94d}
\definecolor{bad}{HTML}{ee312f}
\definecolor{fern}{HTML}{3C60B8}

\title{FEABench: Evaluating Language Models on Multiphysics Reasoning Ability}

\author{Nayantara Mudur$^{2, 1}$\thanks{Work mainly done as a student researcher at Google Research.} \quad Hao Cui$^{1}$ \quad Subhashini Venugopalan$^{1}$  \quad Paul Raccuglia$^{1}$ 
\\ 
\textbf{Michael P. Brenner}$^{1, 2}$ \quad \textbf{Peter Norgaard}$^{1}$ \\
$^{1}$Google Research \quad $^{2}$Harvard University \\
\texttt{\{vsubhashini,praccu,mbrenner,pnorgaard\}@google.com} \\
\texttt{nmudur@g.harvard.edu}
}

\begin{document}

\ifcolmsubmission
\linenumbers
\fi

\maketitle

\begin{abstract}
Building precise simulations of the real world and invoking numerical solvers to answer quantitative problems is an essential requirement in engineering and science. We present FEABench, a benchmark to evaluate the ability of large language models (LLMs) and LLM agents to simulate and solve physics, mathematics and engineering problems using finite element analysis (FEA). We introduce a comprehensive evaluation scheme to investigate the ability of LLMs to solve these problems end-to-end by reasoning over natural language problem descriptions and operating \comsol, an FEA software, to compute the answers. We additionally design a language model agent equipped with the ability to interact with the software through its Application Programming Interface (API), examine its outputs and use tools to improve its solutions over multiple iterations. \rewrite{Our best performing strategy generates executable API calls \change{88\%} of the time.} LLMs that can successfully interact with and operate FEA software to solve problems such as those in our benchmark would push the frontiers of automation in engineering. Acquiring this capability would augment LLMs' reasoning skills with the precision of numerical solvers and advance the development of autonomous systems that can tackle complex problems in the real world.  \ifcolmpreprint The code is available at \href{https://github.com/google/feabench/tree/main}{FEABench} \github.
\fi
\end{abstract}

\section{Introduction}
\added{Several works have demonstrated the significant potential of large language models (LLMs) in scientific and mathematical domains \citep{lewkowycz2022solving, yang2024leandojo,hendrycks2021measuring,rein2023gpqa,trinh2024solving, kumarappan2024leanagent,chung2025theoretical}. However, existing work has largely focused on analytical mathematical and scientific reasoning skills or the ability to generate code in general purpose programming languages \citep{tian2024scicode,jimenez2023swe}. Addressing the degree of complexity required in \rewrite{numerical simulation-intensive} science and engineering workflows – which requires the composition of scientific reasoning with the ability to operate simulation software – remains an outstanding challenge.} Many quantitative tasks that form the cornerstone of these workflows require numerical analysis performed with sophisticated computational modeling software. For example, the development of a smartphone requires detailed modeling of the mechanical, thermal, and electrical behaviors of its many subcomponents. Finite element analysis (FEA) (eg: \cite{courant1994variational}) software develops approximate solutions to the underlying partial differential equations for a physical system, by building discretizations (or meshes) over geometries. The resulting equations are then solved numerically. The vast real-world relevance of FEA to domains like mechanical, biomedical and aerospace engineering, consumer electronics, manufacturing, and scientific research has given rise to software such as \comsol \citep{comsol,multiphysics1998introduction}, that are indispensable to modeling complex systems with the interplay of non-trivial geometries, and multiple physical phenomena.

Despite the potential impact, the application of LLMs to engineering simulation tasks like FEA remains largely unexplored. In this paper, we begin to bridge this gap by \added{measuring the ability of} LLMs and LLM-agents to build models and solve engineering problems using finite element analysis.

Our contributions are the following: 
\begin{itemize}[leftmargin=2em]\itemsep0em
    \item We introduce a benchmark intended for LLM and  agentic research on engineering simulation\added{, a novel domain for LLM benchmarks}. The benchmark consists of (1) \benchmain{}: \change{15} manually verified problems, in addition to (2) \benchlarge{}: a larger set of \change{200} algorithmically parsed tasks. The problems in \benchmain { are} (a) quantitatively \textit{verifiable}, that is, if solved completely and correctly, a desired target value will be computed and exported to a table, (b) manually confirmed to have input problem descriptions that are \textit{self sufficient} and do not omit information necessary to solve the problem (c) manually verified to be \textit{solvable}, i.e. we confirmed that if the steps to model the problem are followed faithfully in \comsol{} the desired target value is computed. The target values are expected to be largely independent of the modeling software. The LLM's objective is to read the problem specification and operate \comsol by generating a sequence of Java calls to its API that would build the model and compute the target. \added{The skills this requires include (1) code generation in a \textit{low-resource} setting,} (2) inferring spatial dimensions and representing objects as compositions of geometrical primitives, (3) making correct and consistent physics reasoning decisions (e.g.: boundary conditions and properties).
    \item We further define two versions of the tasks in \benchmain -- \model and \plan, to probe different versions of task complexity. 
    \item We introduce a holistic automated evaluation strategy with intermediate metrics that seek to \rewrite{measure different facets of `distance to a correct solution'}. We benchmark different SOTA LLMs on their baseline (single-turn) performance with these metrics.
    \item Finally, we design an \cready{interface in which an LLM can interact with the \comsol{} API and with specialized auxiliary functions and and build an agent that uses execution feedback to} improve its solution over multiple turns.
\end{itemize}

We selected \comsol as the framework for our benchmark because it is extensively used for commercial engineering analysis as well as scientific research and supports a wide range of physics models. The FEA workflow is relatively canonical, the reasoning approach for modeling is similar to other FEA software, and problems typically involve a shared conceptual breakdown into a sequence of steps that involve defining 1) Geometry, 2) Material properties, 3) Physics, 4) Mesh 5) Numerical Analysis and Solver settings, and 6) Postprocessing (details in \aref{app: codeblocks}).

\begin{figure}[t]
\centering
\includegraphics[width=\textwidth]{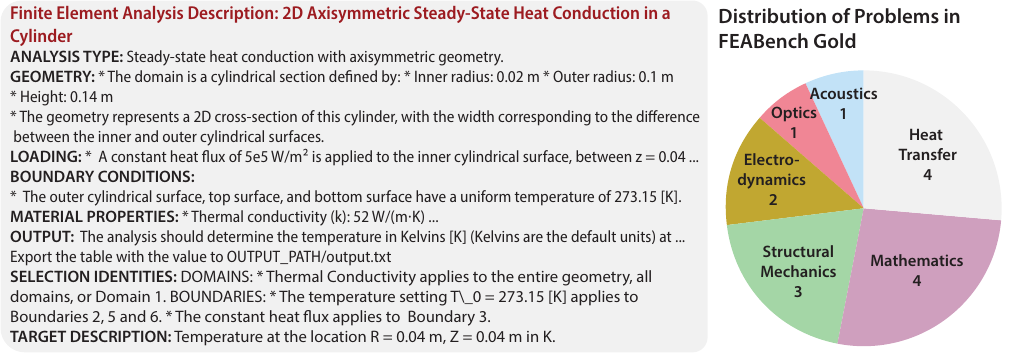}
\caption{\textit{Left:} Illustrative abbreviated example of Model Specifications for one of the heat transfer problems. 
\textit{Right:} Distribution of \benchmain{ problems} by physics domain.}
\label{fig:schema}
\end{figure}

\section{Datasets and Tasks}
\label{sec:data}
\paragraph{\benchmain} The benchmark problems are derived from tutorials in the \cite{tuts} Application Gallery  and are often based on established validation problems or other sources (eg: \cite{melnik2003bandstructures, national1990standard}). The input is a natural language problem description with a specific target quantity that needs to be computed (\fref{fig:schema}). The problems span a range of real world / mathematical systems including heat transfer in objects, modeling a stock option using the Black-Scholes Equation and eigenfrequency analysis of a quantum dot and a beam. Each entry consists of the following main fields:
\begin{itemize}[leftmargin=2em]\setlength\itemsep{-0.2em}
    \item \textbf{Model Specifications:} A complete description of the task, including geometry, material properties, physics specifications, initial/boundary conditions, and the output to be computed. This field is intended to be general enough to be relevant to other softwares or approaches, but is unambiguous about details such as material properties.
    \item \textbf{Selection Information:} An engineer would typically identify spatial information like geometric selections (points, boundaries, and domains) using the Graphical User Interface (GUI). \rewrite{We provide this field as a substitute for images for LLMs and agents without the ability to receive visual input from the GUI.} This information is valid as long as the agent chooses to construct the geometry in \comsol in a manner that is reasonably similar to the construction of the ground truth (GT) geometry.
    \item \textbf{Plan:} Step-by-step instructions to solve the problem using \comsol.
    \item \textbf{Target Description:} A brief phrase describing the quantity that needs to be computed.
    \item \textbf{Target Value:} The correct value of the target physical quantity.
    \item \textbf{Ground Truth Code: } Lines of \comsol API calls that can be executed to build a model that successfully computes the target value.
    \item \textbf{Model Tree: } Executing \comsol calls can be regarded as modifying a tree with certain predefined \textit{branches} such as \textit{geometry} and \textit{physics}. The model generated by executing code can thus be represented in a condensed form as a model tree (see \aref{app: example_bench}). This is a high-level lossy representation of a solution path, as the code cannot be exactly recovered from the model tree.
\end{itemize}
\vspace{-0.3cm}
Converting tutorials to verifiable benchmark problems requires ensuring that an artifact can be computed from it, generating inputs and the GT solution and verifying that it computes the correct target value (\aref{A: Dataset}).

\paragraph{\benchlarge} \rewrite{We further evaluate SOTA LLMs on a larger dataset consisting of \change{200} \cite{tuts} Application Gallery tutorial problems. Since these are algorithmically parsed from tutorials, and most tutorials are for demonstrative purposes, the tasks are not structured so as to export a verifiable numerical artifact. They may instead instruct the user to generate specific plots or compute tables. The input consists of a field termed `\textbf{Plan}', which corresponds to the Modeling Instructions in the tutorial. This specifies explicit instructions (similar in nature to the Plan field in \benchmain). We additionally save the ground truth API calls in `\textbf{Code}' after running some preprocessing steps on the ground truth API calls, in order to resemble the format of the code in \benchmain}.

\paragraph{Annotated Library} We additionally generate a set of 768 annotated code snippets, by querying an LLM (Gemini-1.5-Flash) to translate code blocks to natural language summaries. Unlike the previous two datasets described, we do \textit{not} use this for evaluation. This is used to retrieve snippets in our agent system.
\subsection{\added{Tasks}}
We propose two task variants for LLMs to solve under \benchmain:
(1) the \model task, in which the input description for each problem consists of the Model Specifications and Selection Information fields and, (2) The \plan task, in which the input description consists of the Plan field.
In both cases the LLM agent is expected to return a solution that should consist of the API calls that solve the problem, similar to \textbf{Ground Truth Code}. When executed in the \comsol API during evaluation, a correct solution will export a table containing a computed value that should match the \textbf{Target Value}. We note that both task inputs i.e. the Model Specifications along with the Selection Information, or Plan are independently self-sufficient problem formulations for the LLM to solve using the API. The \textbf{Model Specifications} field most closely resembles a naturally occurring \rewrite{real-world problem description}. We conduct experiments primarily on \benchmain, unless specified otherwise.



\begin{figure}[!ht]
\centering
\includegraphics[width=0.7\textwidth]{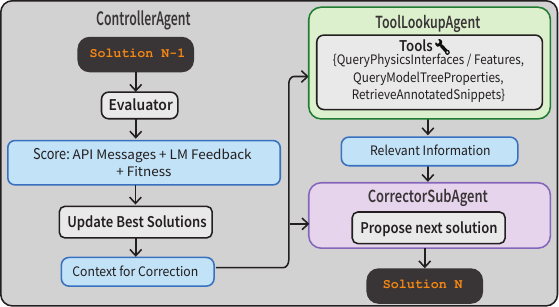}
\caption{An overview of the agent and environment design.}
\label{fig:agent}
\end{figure}
\section{Approach}
We first consider a baseline (non-interactive) setting where the LLM generates code without the ability to execute it. \added{The \comsol API provides several `physics interfaces' that describe different physics systems and phenomena. A user or LLM needs to select the correct interface and create and modify features under it, in order to implement the physics of the problem.} We consider two prompting strategies for both task variants. (1) The first prompting strategy involves a \textbf{One-Shot} prompt where the model is given one full input and solution example. (2) The second strategy, \textbf{PhyDoc In-Context}, includes the list of valid physics interfaces and features under them in addition to the one-shot example. To further address the challenge of correctly operating the simulation software, we build an agentic system that interacts with the software, and uses tools to improve its solution.

\subsection{A Multiphysics Reasoning Agent}
\verbose{Recent work has sought to explore the space of designing optimal Agent-Computer Interfaces \citep{yang2024swe,wang2024executable} primarily for software engineering. However, these frameworks (see \aref{app:related} for a discussion of related work) are tailored to codebase navigation and bash execution: utilities crucial to software development, but of limited relevance \added{to engineering simulation environments.}} We design a multi-agent system that interacts with the \comsol API, as well as tools (or specialized functions). The system is equipped with an \texttt{Evaluator} whose feedback is used to compute a ‘fitness’ for each solution. This is used to track the best solutions. The \texttt{ControllerAgent} calls a \texttt{CorrectorSubAgent} that proposes the next solution given the ‘current’ code and feedback, execution history and the result of tool calls. It in turn delegates tool calls to the \texttt{ToolLookupAgent}. To minimize failures or longer-than-desired chains of calls, we adopt an algorithmic sequence of agent calls \textit{except} within the \texttt{ToolLookupAgent}.  An initial set of 20 samples are generated using PhyDoc In-Context
and the best solutions are corrected for 20 steps (see Appendix C.2 for details). The best of all solutions is then identified and evaluated.
The tool registry consists of a retriever tool \texttt{RetrieveAnnotatedSnippets}, a tool that queries the API to return a dictionary of properties under a specified node \texttt{QueryModelTreeProperties} and two tools that return the list of valid physics interfaces and features (\texttt{QueryPhysicsInterfaces} and \texttt{QueryPhysicsFeatures}). We describe our implementation and selection strategy in \aref{app:agent}, but summarize key features below:
\begin{itemize}[leftmargin=2em]\itemsep0em
\vspace{-4pt}
\item \cready{LLM-Assisted Semantic Code Search: To address the challenge of low-resource code generation, we create a \texttt{RetrieveAnnotatedSnippets} tool that allows the LLM to search the Annotated Library (\sref{sec:data}) for syntactically correct code snippets relevant to a given step (eg: `Define the thermal\ldots' under `material' in \fref{fig:tool_demo}, right panel, \textcircled{\small{3}}).
\begin{figure}[!ht]
\includegraphics[width=\textwidth]{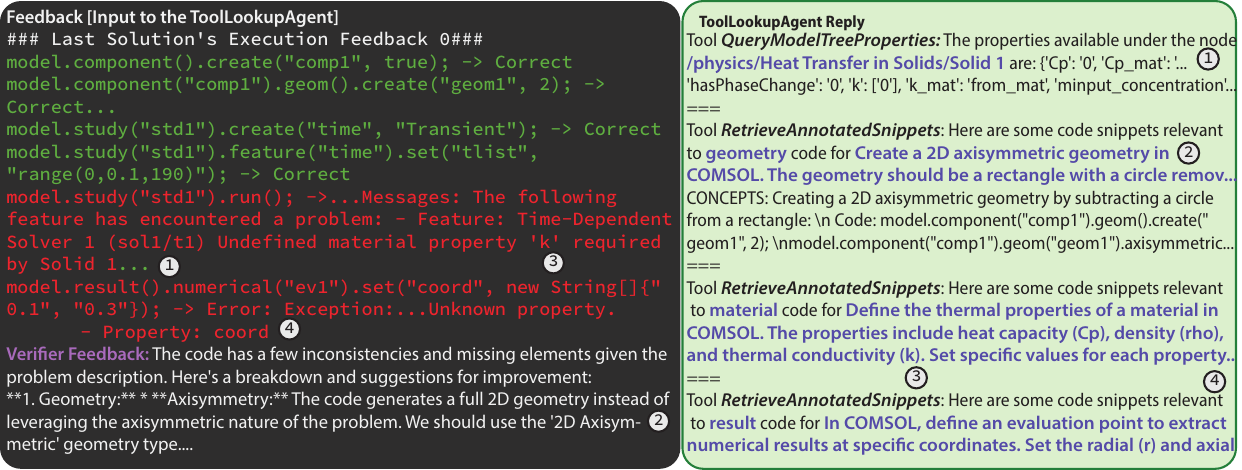}
\caption{The \texttt{Evaluator}'s feedback (\textit{left}) is passed to the \texttt{ToolLookupAgent}, that calls \textbf{\textit{tools}} and returns their concatenated output. {\color{lm} \textbf{Violet}}, on the \textit{left} indicates that the Verifier Feedback is returned by an LLM, and on the \textit{right}, denotes the arguments chosen by the \texttt{ToolLookupAgent} to call the tools with. The numerical annotations highlight the correspondence between the errors and arguments.}
\label{fig:tool_demo}
\end{figure}

\item Hybrid Evaluator Feedback: The API returns a `reply' to each line of code in the parsed LLM solution. \added{These messages indicate whether a line of code was correctly executed. We define the percentage of correctly executed parsed lines of code as `executability'.} API messages alone, however, will not contain information about inconsistencies with the problem description, such as incorrect physical units. Thus, high executability does not guarantee alignment. To address this gap, we call a VerifierLLM to provide feedback (\fref{fig:tool_demo}, left panel) when executability crosses 90\%. The API feedback provides a signal on \textit{syntactical correctness} and the VerifierLLM provides a signal on \textit{alignment} and \textit{completeness}.

\item Analytical-Numerical Consistency: Several problems may allow a scientist to formulate an approximate analytical guess for the target value, even if a precise value may only be derivable numerically. Using this principle, the VerifierLLM sets an analytical guess at the start of the Multi-Turn experiment, given the problem description and compares the numerically computed target with the analytical guess.} 
\end{itemize}




\section{Evaluation Metrics}
\label{sec:eval}
Reasoning correctly about the problem and issuing the right calls to operate the API poses a challenging task for even SOTA LLMs. Moreover, a model can only compute a correct target value if it could generate all the code to solve the problem successfully. This renders conventional execution-based code evaluation metrics such as the `\textit{pass@k}' metric \citep{chen2021evaluating,kulal2019spoc} challenging to apply to this setting, since most solutions are unable to completely solve the problem. \added{Text-based similarity metrics, on the other hand, are confounded by the preponderance of boilerplate code and the functional equivalence of different code blocks (\aref{app:codesim}). To address this challenge,} we introduce a multifaceted evaluation strategy that measures correctness even when a target value could not be computed (\tref{tab:metrics_2}). Metrics denoted by \textsuperscript{\textdagger} require execution of the API calls. We delineate the metrics here:
\vspace{-4pt}
\definecolor{gold}{HTML}{bf9001}
\begin{itemize}[leftmargin=2em] \setlength\itemsep{-0.1em}
\item \textbf{Executability\textsuperscript{\textdagger}:} Executable lines as a fraction of parsed API calls in an LLM solution. A given line may be invalid if it is syntactically incorrect or if it refers to an invalid action (like modifying a property under a non-existent node).

\item \textbf{Model Tree Score\textsuperscript{\textdagger}:} Similarity score between the LLM solution's model tree and a GT tree. This is normalized so that a solution with no parsed lines of code is scored 0. If it was equivalent to the GT tree, the score would be 1. This measures the \textit{alignment} of the model's solution path with a successful path.

\item \textbf{Physics Metrics:} The metrics above analyzed the \textit{entire} solution or its derived artifacts. The code is a basis to represent the actions the LLM takes to model the problem. Since the physics block is both the most diverse and the most challenging (\fref{fig:blockwise}), we further evaluated specifically the LLM's physics actions. The most basic physics action sequence involves: Create Interface (e.g.: \texttt{HeatTransfer}) $\rightarrow$ Create Feature under Interface (e.g.: \texttt{TemperatureBoundary}) $\rightarrow$ Modify Feature Properties (e.g.: \texttt{T0}, to set a temperature). Our Physics Metrics include (a) \textit{Interface Factuality}: What fraction of interfaces created by the LLM are real \comsol interfaces and \textit{not} hallucinated? (b) \textit{Interface / Feature / Feature Property Recall}: How many interfaces / features / feature properties created / modified by the GT solution were also in the LLM solution? (c) \textit{Feature Dimension:} For features created by both, does the feature's spatial dimension match? For example, if an LLM sets a temperature boundary condition on a 1D geometry, this metric checks whether it deduced that the boundary should be 0 dimensional, by comparing the dimension with the GT boundary dimension. \cready{While these metrics offer a granular look into the LLM's physics reasoning path, some nested physics metrics, such as `Feature Dimension' will not be valid for a problem when there is no overlap between the GT and the LLM code: we mask out these problems while computing the means for that metric.}
\item \textbf{Target Relative Error\textsuperscript{\textdagger}}: We entask an LLM 
to check that the computed value in the exported table matches the target description and is not a default value, and to parse the response, if so. \textbf{Valid Target} is the number of problems for which the LLM judges the value to be valid. We then compute the relative error between the value and the Target (GT) Value. \textbf{{\color{gold}Relative Error $\vert$ Strict}} computes the mean relative error only over problems for which Valid Target is True, \textsc{AND} the relative error is less than 10\%. \textit{Relative Error $\vert$ Strict is the principal metric one would ideally use to assess whether a problem was truly solved.}
\end{itemize}
The tables report the means and standard errors on the mean across problems that the experiments were run on. Some nested physics metrics, such as `Feature Dimension' might not be valid for a specific problem, if there was no matching feature between the GT and the LLM code: we mask out these problems while computing the means for that specific metric.
\begin{table}
\centering 
\begin{small}
\caption{Summary of Evaluation Metrics} 
\label{tab:metrics_2}
\begin{tabular}{lcccc}  
\toprule
\textsc{Metric} & \textsc{Artifacts} & \multicolumn{3}{c}{\textsc{Skills Measured}} \\
\cmidrule{3-5}
 &  & Correctness & Alignment & Physics Reasoning \\
\midrule  
Executability & API Messages & $\checkmark$ &  &   \\
Model Tree Score & Model Tree &  & $\checkmark$ & \\
\midrule 
Physics Metrics & Physics Code &  &  &  \\
\hspace{.5cm}Interface Factuality &  & $\checkmark$ &  &  \\ 
\hspace{.5cm}Recall Metrics &  &  & $\checkmark$ & $\checkmark$ \\
\hspace{.5cm}Feature Dimension &   & $\checkmark$ &  & $\checkmark$ \\
\midrule 
Target Value Metrics & Output &  $\checkmark$  & $\checkmark$ & $\checkmark$ \\
\bottomrule
\end{tabular}
\end{small}
\end{table}
\section{Results and Discussion}
\textbf{Comparison across LLMs at baseline.}
\definecolor{gray}{gray}{0.5}
\newcommand\omitifnospace[1]{{\color{black}#1}}
\change{Three closed-source LLMs -- Claude-3.5-Sonnet \citep{claude}, GPT-4o \citep{gpt4orelease} and Gemini-1.5-Pro \citep{reid2024gemini}} -- and three open-weights LLMs from the Gemma family \citep{team2024codegemma,team2024gemma} are tested on the \model task \ifthenelse{\boolean{owa}}{
}{
under \benchmain}, with a one-shot prompt (\tref{tab:compare_models} and \ref{tab:compare_models2}). We find that closed-weight models are able to generate code with moderate executability \change{$\sim0.60-0.79$}, implying that LLMs are familiar with the higher-level grammar and syntax of \comsol API calls or can infer it from the one-shot example. Getting more granular choices correct is more challenging: LLMs are prone to hallucinating the interface choice (factuality between \cready{[0.54-0.85]}). The open-weights LLMs generally perform worse, especially on the alignment-probing metrics like the Model Tree Score and Physics Recall\footnote{Since these LLMs were unable to define matching features (feature recall), the dimension metric could be evaluated for fewer than 5 problems and was thus omitted in \tref{tab:compare_models2}.}. We also compare the performance of the closed-source LLMs on the 200 problems in \benchlarge. Unlike the human-verified \benchmain, \benchlarge { instances} do not have a single final target artifact, so we only evaluate these against metrics  

\label{tab:compare_models}
\begin{table}[!h]
\begin{small}
\centering
\caption{Code Metrics: Comparison on \model across LLMs.}
\label{tab:compare_models}
\begin{tabular}{p{1.2in}p{.7in}p{1in}p{1in}}
\toprule
Experiment & Executability & Model Tree Score & Valid Target \\
\midrule
Claude 3.5 Sonnet & \textbf{0.79}$\pm$0.03 & \textbf{0.69}$\pm$0.07 & \textbf{1}/15 \\
GPT-4o & 0.78$\pm$0.03 & 0.56$\pm$0.06 & 0/15 \\
Gemini-1.5-Pro & 0.60$\pm$0.05 & 0.46$\pm$0.07 & 0/15 \\
\midrule
Gemma-2-27B-IT & 0.56$\pm$0.05 & 0.47$\pm$0.07 & 0/15 \\
Gemma-2-9B-IT & 0.44$\pm$0.06 & 0.28$\pm$0.06 & 0/15 \\
CodeGemma-7B-IT & 0.52$\pm$0.07 & 0.35$\pm$0.06 & 0/15 \\
\bottomrule
\end{tabular}
\vspace{0.1cm}
\caption{Physics Metrics: Comparison on \model across LLMs.}
\label{tab:compare_models2}
\begin{tabular}{p{1.1in}p{.75in}p{.6in}p{.6in}p{.7in}p{.7in}}
\toprule
Experiment & Interface \mbox{Factuality} & Interface Recall & Feature \mbox{Recall} & Feature Property Recall & Feature Dimension \\
\midrule
Claude 3.5 Sonnet & \textbf{0.85}$\pm$0.10 & \textbf{0.71}$\pm$0.13 & \textbf{0.80}$\pm$0.10 & \textbf{0.22}$\pm$0.10 & \textbf{0.95}$\pm$0.05 \\
GPT-4o & 0.79$\pm$0.11 & 0.64$\pm$0.13 & 0.55$\pm$0.12 & \textbf{0.22}$\pm$0.11 & \textbf{0.95}$\pm$0.05 \\
Gemini-1.5-Pro & 0.54$\pm$0.14 & 0.43$\pm$0.14 & 0.39$\pm$0.10 & 0.15$\pm$0.09 & 0.86$\pm$0.14 \\
\midrule
Gemma-2-27B-IT & 0.69$\pm$0.13 & 0.50$\pm$0.14 & 0.14$\pm$0.08 & 0.11$\pm$0.07 & \leavevmode{\color{gray} -} \\
Gemma-2-9B-IT & 0.70$\pm$0.15 & 0.43$\pm$0.14 & 0.06$\pm$0.04 & 0.07$\pm$0.07 & \leavevmode{\color{gray} -} \\
CodeGemma-7B-IT & 0.45$\pm$0.13 & 0.21$\pm$0.11 & 0.17$\pm$0.09 & 0.07$\pm$0.07 & \leavevmode{\color{gray} -} \\
\bottomrule
\end{tabular}
\end{small}
\end{table}

\begin{table}[!h]
\begin{small}
\centering
\caption{Comparison across models on FEABench Large.}
\label{tab:large}
\begin{tabular}{p{1in}p{.6in}p{.52in}p{.65in}p{.6in}p{.6in}}
\toprule
Experiment & Interface Factuality & Interface Recall & Feature \mbox{Recall} & Feature Property Recall & Feature Dim. \\
\midrule
Claude 3.5 Sonnet & \textbf{0.68}$\pm$0.03 & \textbf{0.50}$\pm$0.03 & \textbf{0.49}$\pm$0.03 & \textbf{0.29}$\pm$0.02 & \textbf{0.96}$\pm$0.01 \\
GPT-4o & 0.66$\pm$0.03 & 0.48$\pm$0.03 & 0.26$\pm$0.03 & 0.20$\pm$0.02 & 0.82$\pm$0.05 \\
Gemini-1.5-Pro & 0.57$\pm$0.04 & 0.28$\pm$0.03 & 0.44$\pm$0.03 & 0.20$\pm$0.02 & 0.72$\pm$0.04 \\
\bottomrule
\end{tabular}
\end{small}
\end{table}
that don't require execution. \textit{Claude 3.5-Sonnet consistently has the best performance across metrics on both benchmarks.}

\subsection{Analyzing factors that contribute to complexity}
To analyze sources of difficulty in a single-query, non-interactive setting, first, we use the two task versions to decouple whether the bottleneck lies in making correct reasoning decisions or in translating explicit natural language steps into syntactically correct code. \added{Next, we examine gains from changing the prompting strategy. Finally,} we leverage the common structure of the code across problems to examine which block is the most challenging. We fix the LLM to Gemini-1.5-Pro in subsequent \comsol-based experiments. We include a detailed qualitative analysis of a single solution in \aref{app:qual}.

\begin{figure}
    \centering
    \includegraphics[width=0.4\textwidth]{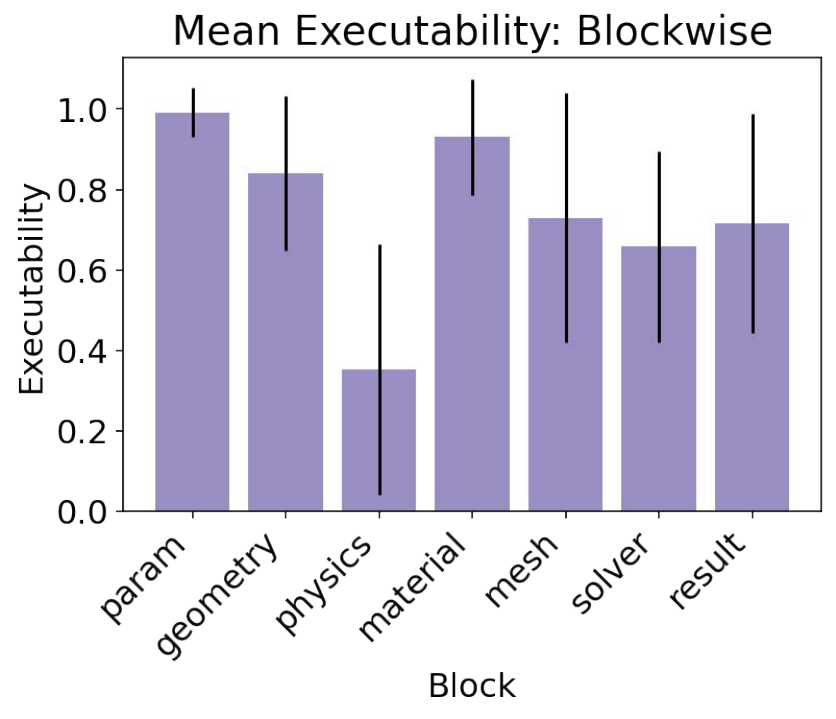}
\caption{Block-wise executability across 300 samples of code and Gemini-1.5-Pro. The physics block has the lowest executability. Error bars denote standard \mbox{deviations}.}
\label{fig:blockwise}
\end{figure}

\paragraph{LLMs find it challenging to translate physics reasoning decisions to code.}
We examine whether the \plan task is easier than the \model task. The comparison between task versions is of interest since both demand slightly different skills. \model requires the \textit{composition} of planning and reasoning about engineering decisions with translation to valid API calls. Eg: In \fref{fig:schema}, the LLM needs to infer that the correct representation of a cylinder's 2D cross-section is a rectangle. The \plan { task} explicitly describes all steps to be followed in natural language and requires the LLM to only \textit{translate} them to valid calls. \rewrite{The comparison between the two tasks offers a way to decouple the difficulty arising from making correct modelling decisions from translating the decisions into calls with the correct syntax. If an LLM or a user fared poorly at making correct modelling decisions but could reliably translate natural language instructions to API calls, it would find \plan an easier task.} However, we find that \textit{providing an explicit plan doesn't consistently boost performance on \benchmain}. We hypothesize this could be due to  the LLM hallucinating API calls by following natural language instructions verbatim. \change{For instance, for Heat Transfer problems, the natural language instructions in \plan instruct the LLM to construct a `Heat Transfer in Solids' interface. However, the syntactically correct interface name is \texttt{HeatTransfer}. This is also observable in the slight drop on Interface Factuality between the two tasks in \tref{tab:compare_ongem2}.}
\paragraph{Grounding the LLM with API information boosts performance.} The comparison between task formulations indicates that correctly translating decisions to code is a larger bottleneck for our dataset than making correct decisions. We now assess performance, with the list of physics interfaces and features included in the prompt (PhyDoc In-Context). This helps performance, particularly reducing interface hallucinations \change{(factuality: \model: 0.54$\rightarrow$1.0, \plan: 0.38$\rightarrow$0.85)}.
\paragraph{Physics specific blocks are the most challenging.}
\fref{fig:blockwise} analyzes executability across LLM solutions by breaking down line-wise executability by the block of code the line belongs to. We used the \added{initial set of} 20 samples for the 15 problems with the PhyDoc In-Context prompt \added{from the Agent experiment}. \textit{The physics block has the lowest executability} with a single query. This motivates our focus on evaluation metrics that focus on the physics block and tools that seek to help ground the LLM's code with physics-specific information.

\begin{table}[tp]
\begin{small}
\centering
\caption{Code Metrics: Comparison across tasks, prompts and agents.}
\label{tab:compare_ongem}
\begin{tabular}{p{2in}p{.7in}p{1in}p{.7in}}
\toprule
Experiment & Executability & Model Tree Score & Valid Target \\
\midrule
\model: One-Shot & 0.60$\pm$0.05 & 0.46$\pm$0.07 & 0/15 \\
\model: PhyDoc In-Context & 0.62$\pm$0.05 & 0.58$\pm$0.07 & 1/15 \\
\model: Multi-Turn Agent & \textbf{0.88}$\pm$0.03 & 0.56$\pm$0.08 & \textbf{2}/15 \\
\midrule
\plan: One-Shot & 0.54$\pm$0.03 & 0.39$\pm$0.03 & 0/15 \\
\plan: PhyDoc In-Context & 0.59$\pm$0.05 & \textbf{0.59}$\pm$0.06 & 0/15 \\
\bottomrule
\end{tabular}
\vspace{0.1cm}
\caption{Physics Metrics: Comparison across tasks, prompts and agents.}
\label{tab:compare_ongem2}
\begin{tabular}{p{1.86in}p{.58in}p{.45in}p{.45in}p{.5in}p{.45in}}
\toprule
Experiment & Interface Factuality & Interface Recall & Feature Recall & Feature Property Recall & Feature Dim. \\
\midrule
\model: One-Shot & 0.54$\pm$0.14 & 0.43$\pm$0.14 & 0.39$\pm$0.10 & 0.15$\pm$0.09 & 0.86$\pm$0.14 \\
\model: PhyDoc In-Context & \textbf{1.00}$\pm$0.00 & 0.71$\pm$0.13 & 0.48$\pm$0.10 & 0.08$\pm$0.07 & 0.59$\pm$0.16 \\
\model: Multi-Turn Agent & 0.93$\pm$0.07 & \textbf{0.79}$\pm$0.11 & \textbf{0.75}$\pm$0.09 & 0.24$\pm$0.10 & 0.89$\pm$0.07 \\
\midrule
\plan: One-Shot & 0.38$\pm$0.14 & 0.36$\pm$0.13 & 0.43$\pm$0.11 & \textbf{0.32}$\pm$0.11 & 0.79$\pm$0.15 \\
\plan: PhyDoc In-Context & 0.85$\pm$0.10 & 0.57$\pm$0.14 & 0.47$\pm$0.11 & 0.13$\pm$0.07 & \textbf{0.93}$\pm$0.07 \\
\bottomrule
\end{tabular}
\end{small}
\end{table}
\newpage
\added{\subsection{Agent Results}}
\added{Our results underscored the need to ground the LLM's responses with feedback from and documentation about the API.} The interactive Multi-Turn Agent has the best performance of all the \comsol experiments on the \model task on several metrics including executability \change{(0.62 $\rightarrow$0.88)}. Although \textbf{Relative Error $|$ Strict} is the principal metric one would ideally optimize for, we do not report means over that metric here since the LLM was only able to pass the `Strict' cut (i.e. compute a valid target that was also within 10\% of the correct answer) \change{for a single problem in the Multi-Turn Agent and ModelSpecs + PhyDoc In-Context experiments}. For this problem, the correct target value is 18.3$^\circ$ Celsius, and the value exported by the LLM is 20$^\circ$ Celsius (specifically 19.999...$^\circ$ Celsius), which is a default temperature in \comsol: this is an indicator of the solution not being solved correctly. \rewrite{While a stricter relative error threshold would eliminate such serendipitous matches, this risks filtering out problems in which a solution might be conceptually correct but differs from the target because of say, differences in numerical choices.}

\added{In \aref{app:python}, we additionally examine how well the problems can be solved in Python by `\texttt{SWE-agent}', a software-engineering agentic framework \citep{yang2024swe}. In this setting, generating executable code is no longer a challenge. However, the bottleneck now lies in achieving the desired precision, due to the absence of verified physics modules and automatic numerical solvers in Python, which requires a user / LLM to explicitly define all equations from scratch. The agent is able to compute a valid target for 11 problems but is only able to compute a solution that passes the `Strict' cut for 4 problems.}

\section{Conclusion and future directions}
\added{FEABench addresses a key gap in the application of LLMs to scientific disciplines by analyzing their ability to operate engineering simulation software to solve problems that require numerical analysis to model mathematical and physical systems. The combination of capabilities required include low-resource code generation and physics and spatial reasoning skills. The complex compositional nature of this task makes it a novel }testbed to measure the ability of agentic approaches to interact with a simulation environment and master a domain-specific language well enough to solve real-world quantitative problems. \added{We designed a multiphysics reasoning agent with specialized tools and hybrid feedback to enhance the ability of LLMs to generate executable code. By introducing a multifaceted evaluation strategy and different task formulations, we analyzed the bottlenecks to succeeding at the tasks. }Addressing these challenges would advance the development of agentic systems for engineering modelling and simulation.

A way to further increase the complexity of the benchmark could involve more intricate geometries, imported Computer-Aided Design (CAD) models, and requiring the LLM to operate the software via its graphical user interface. \rewrite{While datasets such as \benchlarge{ provide} a useful statistical signal on the quality of code solutions generated across a large number of problems, adding more human verified problems would be valuable. Using an LLM-annotated corpus to boost code executability might facilitate code generation in other low-resource domain-specific language contexts. Conversely, code generation approaches for other low-resource languages \citep{cassano2024knowledge} might reduce the bottleneck of translating predefined decisions to code. 


The ability to operate engineering simulation software to quantitatively analyze a problem would augment LLMs' reasoning skills with the software's numerical precision and inbuilt checks, and significantly push the ceiling on the complexity of problems that LLMs can accurately and reliably solve. Unlocking this potential would bring LLMs a step closer to being able to serve as grounded `engineering assistants' that can autonomously run precise simulations to innovate and optimize designs and answer quantitative questions about physical phenomena in the real world.}

\section{Reproducibility Statement}
\ifcolmsubmission 
\added{We have included the benchmark problems for { \benchmain} along with the code for the LLM agents, evaluation, API interface, inference and preprocessing of ground truth API calls with this submission.} 
The code will additionally be provided as a Github link upon acceptance.
\fi
\ifcolmpreprint 
The benchmark problems for \benchmain { and} the code are available at \href{https://github.com/google/feabench/tree/main}{FEABench} \github.
\fi
The prompts used are in \aref{app: prompts}. We will also release the list of tutorial identifiers used in our evaluation on \benchlarge, and the library of code block annotations used in the RetrieveAnnotatedSnippets tool. A \comsol license will be needed to run the Multi-Turn Agent experiment, and to compute execution-based metrics (delineated in Section 3 by $^\dagger$). The bridge to communicate with \comsol from Python is described in \aref{app:mph} and the Python packages needed are open-source. The tutorial documents and models used in \benchlarge{ are} accessible on the internet on the \comsol website.

\ifcolmpreprint
\section{Acknowledgements}
We are grateful to Eser Aygün for valuable suggestions on agent design. We thank Stephan Hoyer and Marc Coram for useful comments on the draft, and are grateful to Rachel Stigler for guidance. At Harvard, NM is partially supported by the National Science Foundation under Cooperative Agreement PHY2019786 (The NSF AI Institute for Artificial Intelligence and Fundamental Interactions).
\fi

\bibliography{references}
\bibliographystyle{colm2025_conference}

\appendix
\section*{Appendix}
\section{Related Work}
\label{app:related}
\paragraph{LLMs and Agents for Code} \cready{Several studies have focused on benchmarking coding in general-purpose programming languages, with a particular focus on software engineering tasks \citep{austin2021program,chen2021evaluating, jimenez2023swe,li2022competition}, and less commonly, science problems \citep{tian2024scicode}.} FEA software emerged because simulating and numerically solving real-world problems from scratch in mainstream languages would require significantly more effort without specialized packages, automatic mesh generation and pre-verified physics modules. \cready{Other work in the LLM literature has focused on optimizing agent-tool call and design such as the ReAct and CodeAct strategies \citep{wang2024executable,yao2022react}. It would be valuable to port blocks from our agent such as the Evaluator and the specialized functions into generalist agentic frameworks like AutoGPT and LangChain \citep{Significant_Gravitas_AutoGPT,Chase_LangChain_2022} to explore possible performance gains and understand the optimum way to distil visual information from the Graphical User Interface (GUI). Beyond the realm of general-purpose programming, some works have sought to incorporate productivity APIs such as those for weather, email among others into agentic workflows \citep{qin2023toolllm,basu2024nestful}. Our agentic approach shares similarities with the Reflexion strategy \citep{shinn2024reflexion}, although in our case the Evaluator mainly subjective feedback from the API, and only queries its VerifierLLM when executability is already high.}
\paragraph{LLMs for Science} The utility of LLMs in science has been explored by evaluating their performance on tasks in medicine \citep{saab2024capabilities,yang2024advancing}, theorem proving \citep{yang2024leandojo}, examination problems of varying levels of difficulty \citep{hendrycks2021measuring, wang2024examining,lewkowycz2022solving} and
 in specific domains such as physics and chemistry \citep{pan2024quantum,bran2023chemcrow}. More recently, there have been efforts to examine whether LLMs can be of utility in other aspects of the scientific process, such as developing hypotheses, reproducibility of code and question-answering \citep{pramanick2024spiqa,mishra2024paperclip,siegel2024core}. \cite{ni2024mechagents} and \cite{tian2024optimizing} made a preliminary exploration on getting LLMs to solve elasticity problems and in a human-in-the-loop setting and \cite{kumar2023mycrunchgpt} explored the role of LLMs on optimizing airfoils.

\section{Dataset Curation}
\label{A: Dataset}
\subsection{\benchmain}
\subsubsection{Selection Criteria:} We chose tutorials that satisfied the following considerations: 
\begin{enumerate}
\item \textit{Simpler Geometry:} \comsol can be used to analyze the physics of systems involving intricate geometries such as microwaves or transformers. In these cases, in practice, most problems involve importing a pre-built geometry object that might have been built externally using Computer-Aided Design (CAD) software and to then perform the remaining analysis. Since we wanted to explore the ability to solve the problem end-to-end and without requiring imports of derived objects, we restrict ourselves to problems that did not require imports of geometry, or any other files. 
\item \textit{Tutorial / Code Simplicity: }We additionally chose problems that did not involve multiple `\texttt{Model}' JAVA classes and restricted ourselves to tutorial documents with fewer than 20 pages. The first requirement is a consequence of how our connection to the \comsol sandbox is set up, and to make the problem easier for the models to attempt to solve. We additionally ensured that the problems were amenable to computing a numerical artifact.
\item \textit{Solving Speed: } We also excluded any problems whose ground truth code took over a minute to solve.
\end{enumerate}

\subsubsection{Generation Procedure:} Without any modification, the tutorials might export a single value, a table, or not export any target quantity at all, with the final output being qualitative in nature, such as in the form of plots or figures. For our benchmark, however, we specifically wanted every problem to have a numerically verifiable target value, in order for there to be an absolute notion of correctness (i.e. if the code was fully correct, and aligned with the intent of the problem, it should be able to export this value). This also enables easier evaluation of the problems. The following procedure and guidelines were adopted to curate the benchmark: \begin{itemize}\itemsep0em
    \item For an initial set of 2-3 problems, model specifications and plans were annotated by hand, by an expert user of \comsol.
    \item For subsequent problems, we speed up the benchmark generation procedure by following an initial LLM-assisted data generation process, with the final verification steps involving humans. An LLM is provided with a tutorial, as well as a two-shot prompt with the expert annotated model specifications.
    \item The LLM is entasked with returning a model specification for the tutorial that has the same format. This requires the LLM to identify an appropriate target value from the tutorial which it does from either the text or the figures, and returning a model specification for computing this target value.
    \item The LLM is then asked to create a plan corresponding to the model specifications, using a two-shot prompt with two plans. The utility of the tutorials are that the plan is closest to the GUI instructions listed in the tutorial, while model specifications is more concise.
    \item A ground truth code that can compute the correct value is then generated for the problem. We manually verify that the ground code when run, exports the desired target value. This step also involves simultaneously ensuring that all information required to build the model is contained in the plan, and in the model specifications by editing the LLM-generated drafts and ensuring that no Translation Errors are encountered when parsing and executing the ground truth code in \comsol using the bridge described in \aref{app:mph} or that any errors if encountered are in non-crucial lines and do not prevent the solution from being computed. Any missing or incorrect information is fixed, and the selection\_information field, that contains numerical identities of boundaries and points is also created.
    \item We add an instruction to export the output to \texttt{OUTPUT\_PATH/output.txt} in the model specifications and plan.
\end{itemize}

\definecolor{framecolor}{HTML}{44485f}
\definecolor{creamywhite}{HTML}{f5f5f5}
\lstnewenvironment{clisting}{\lstset{basicstyle=\ttfamily\small,
breaklines=true}}{}
\lstnewenvironment{plclisting}{\lstset{basicstyle=\rmfamily\bf,
breaklines=true}}{}

\subsubsection{Fields from an example entry:}
\label{app: example_bench}
\begin{figure}
    \centering
    \includegraphics[width=0.8\textwidth]{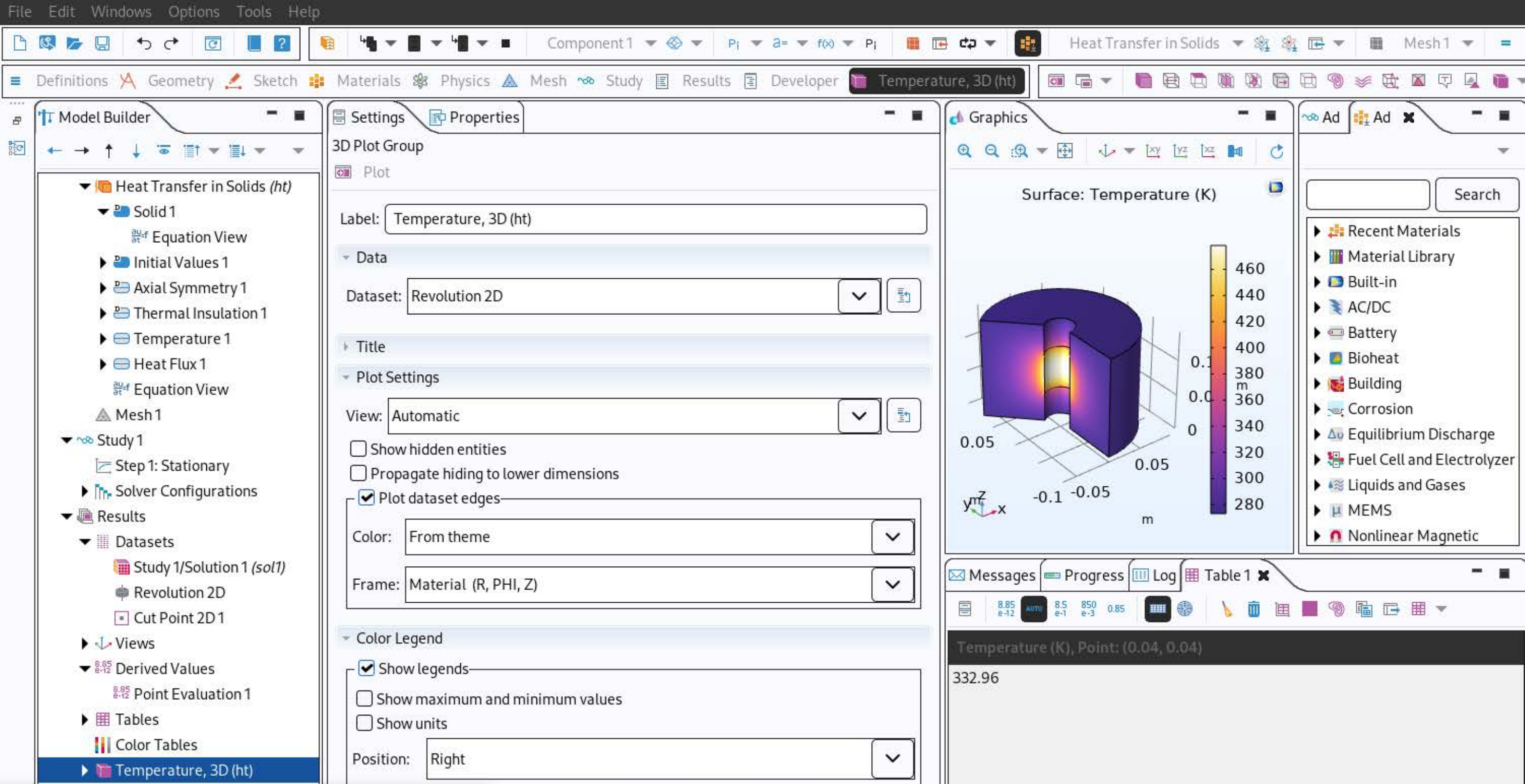}
\caption{Screenshot of the graphical user interface for the correctly solved problem in \fref{fig:schema}.}
\label{fig:comsol_pic}
\end{figure}

Here is an example of the information saved for one of the problems, \texttt{comsol\_453} based on Steady-State 2D Axisymmetric Heat Transfer with Conduction, Heat Transfer Module Application Gallery, \comsol v.6.1. COMSOL AB, Stockholm, Sweden, 2023 \citep{bench453}:\newline
\textbf{Model Specifications: }
\begin{tcolorbox}[width=\textwidth,colback=creamywhite, colframe=black, breakable]

\#\# Finite Element Analysis Description: 2D Axisymmetric Steady-State Heat Conduction in a Cylinder

**ANALYSIS TYPE:** Steady-state heat conduction with axisymmetric geometry.

**GEOMETRY:**
* The domain is a cylindrical section defined by:
\begin{itemize}
    \item Inner radius: 0.02 m
    \item Outer radius: 0.1 m
    \item Height: 0.14 m
\end{itemize}
* The geometry represents a 2D cross-section of this cylinder, with the width corresponding to the difference between the inner and outer cylindrical surfaces.

**LOADING:**
*  A constant heat flux of 5e5 W/m$^2$ is applied to the inner cylindrical surface, between z = 0.04 m and z= 1 m. The remaining portion of the inner cylindrical surface is insulated.

**BOUNDARY CONDITIONS:**
*  The outer cylindrical surface, top surface, and bottom surface have a uniform temperature of 273.15 [K].

**MATERIAL PROPERTIES:**
* Thermal conductivity (k): 52 W/(m·K)

**ELEMENT TYPES:**  The analysis can utilize 2D axisymmetric heat transfer elements.

**MESHES:**  The default mesh can be used.

**OUTPUT:**  The analysis should determine the temperature in Kelvins [K] (Kelvins are the default units) at a specific point on the inner cylindrical surface:

*   Radial Coordinate (r): 0.04 m
*   Axial Coordinate (z): 0.04 m

Export the table with the value to OUTPUT\_PATH/output.txt

\end{tcolorbox}

\textbf{Plan: }
\begin{tcolorbox}[width=\textwidth,colback=creamywhite, colframe=black, breakable]
\#\# Implementing the 2D Axisymmetric Steady-State Heat Conduction in a Cylinder in COMSOL Multiphysics:

**1. Model Setup:**

* **New Model:** Start COMSOL Multiphysics and create a new model.

* **Space Dimension:** Select "2D Axisymmetric".

* **Physics Interface:** Select "Heat Transfer $>$ Heat Transfer in Solids (ht)".

* **Study Type:**  Choose "General Studies $>$ Stationary".
\vspace{0.2cm}

**2. Geometry Definition:**

* **Rectangle:** Create a rectangle representing the cross-section of the cylinder:

    * Width: 0.08 m
    
    * Height: 0.14 m
    
    * Corner Position: (r, z) = (0.02, 0) m
    
* **Point:**

  * In the r field, type 0.02 0.02
  
  * In the z field, type 0.04 0.1
\vspace{0.2cm}

**3. Definitions:**
* **Boundaries:** Define selections for the following boundaries:

    * **Inner Cylinder Surface:**  Left edge of the rectangle
    
    * **Outer Cylinder Surface:** Right edge of the rectangle
    
    * **Top Surface:** Top edge of the rectangle
    
    * **Bottom Surface:** Bottom edge of the rectangle
\vspace{0.2cm}

**4. Physics Settings:**
* **Heat Conduction, Solid:**

    * For the Thermal Conductivity (k), choose User defined, and type 52 W/(m.K).
    
    * Under Thermodynamics Solid, choose User defined for Cp and rho.
    
    * Domain Selection:** Select all domains or Domain 1.

* **Boundary Conditions:**

    * **Temperature:**
    
      * In the Temperature section, type 273.15 [K] for T$_0$.
      
      * Select Boundaries 2, 5 and 6.

    * **Heat Flux:**
    
      * Apply a "Heat Flux" boundary condition with a constant value q0 of 5e5 W/m$^2$.
      
      * Select Boundary 3.
\vspace{0.2cm}

**5. Meshing:**

* **Mesh Creation:** Use the default mesh.
\vspace{0.2cm}

**7. Study Settings:**

* **Solver Configuration:** Use default solver settings for the "Stationary" study.
\vspace{0.2cm}

**8. Analyzing Results:**

* **Temperature at Target Point:**

    * Create a "Cut Point 2D" dataset at this location first and then use that dataset in the point evaluation:
    
      * Locate the Point Data section under Cut Point 2D and type R = 0.04 m, Z = 0.04 m
      
    * Use a "Point Evaluation" feature to evaluate the temperature (in K) at the target point.
    
    * Export the table containing this value to OUTPUT\_PATH/output.txt.
\end{tcolorbox}

\textbf{Selection Information:}
\begin{tcolorbox}[width=\textwidth,colback=creamywhite, colframe=black, breakable]
DOMAINS:
Thermal Conductivity applies to the entire geometry, all domains, or Domain 1.

BOUNDARIES:
* The temperature setting T\_0 = 273.15 [K] applies to Boundaries 2, 5 and 6.

* The constant heat flux applies to  Boundary 3.
\end{tcolorbox}

\textbf{Target Description: }
Temperature at the location R = 0.04 m, Z = 0.04 m in K.

\textbf{Target Value: } 333

\textbf{Target Units: } K

\textbf{Ground Truth Code:}
\begin{tcolorbox}[width=\textwidth,colback=creamywhite, colframe=black, breakable]
\begin{clisting}
model.component().create("comp1", true);

model.component("comp1").geom().create("geom1", 2);
model.component("comp1").geom("geom1").axisymmetric(true);

...
model.component("comp1").physics().create("ht", "HeatTransfer", "geom1");
...
model.component("comp1").physics("ht").create("temp1", "TemperatureBoundary", 1);
model.component("comp1").physics("ht").feature("temp1").set("T0", "273.15[K]");
...
model.result().table("tbl1").comments("Point Evaluation 1");
model.result().numerical("pev1").set("table", "tbl1");
model.result().numerical("pev1").setResult();
model.result().table("tbl1").save("OUTPUT_PATH/output.txt");
\end{clisting}
\end{tcolorbox}

\textbf{Model Tree:}
\begin{tcolorbox}[width=\textwidth,colback=creamywhite, colframe=black, breakable]
\begin{verbatim}
model
├─ parameters
│  └─ Parameters 1
├─ functions
│  ├─ Analytic
│  ├─ Analytic
│  └─ Blackbody Radiation Intensity
├─ components
│  └─ Component 1
├─ geometries
│  └─ Geometry 1
│     ├─ Rectangle 1
│     ├─ Point 1
│     └─ Form Union
...
├─ physics
│  └─ Heat Transfer in Solids
│     ├─ Solid 1
│     │  └─ Opacity 1
│     ├─ Initial Values 1
│     ├─ Axial Symmetry 1
│     ├─ Thermal Insulation 1
│     ├─ Isothermal Domain Interface 1
│     │  └─ Layer Opacity 1
│     ├─ Local Thermal Nonequilibrium Boundary 1
│     ├─ Opaque Surface 1
│     ├─ Continuity 1
│     ├─ Temperature 1
│     └─ Heat Flux 1
...
├─ studies
│  └─ Study 1
│     └─ Stationary
├─ solutions
│  └─ Solution 1
│     ├─ Compile Equations: Stationary
│     ├─ Dependent Variables 1
│     │  └─ Temperature (comp1.T)
│     └─ Stationary Solver 1
│        ├─ Direct
│        ├─ Advanced
│        ├─ Fully Coupled 1
│        ├─ Direct, heat transfer variables (ht)
│        └─ AMG, heat transfer variables (ht)
│           └─ Incomplete LU
├─ batches
├─ datasets
│  ├─ Study 1//Solution 1
│  └─ Cut Point 2D 1
├─ evaluations
│  └─ Point Evaluation 1
├─ tables
│  └─ Table 1
├─ plots
└─ exports
\end{verbatim}
\end{tcolorbox}

\subsection{\benchlarge}
\label{app:large}
The input field in \benchlarge { is} the `Modeling Instructions' section of the tutorial. The output field is the code in the first run function of the exported Java file of the built \comsol model with the following postprocessing steps applied: we append to the last line of each `study' code block in the model with a \texttt{model.study("study\_tag").run();} where "study\_tag" will typically be ``std1'' or ``std2'', and remove the block of `solver' code. While the choice of including the code only in the first run function might make the mapping between instructions and lines of code less one to one in problems with more than one run function, this choice makes this dataset and the style of code resemble the constraints in \benchmain. We make the `study / solver' changes because the `model.sol' code consists of a larger block of automatically populated lines that bear little resemblance to no resemblance to the original problem specification, and often correspond to a single `Compute' step in the GUI. Adding the `.run();' line prompts \comsol to use its default solver best configured to solve the problem depending on the physics and nature of the analysis performed. This is also a pattern guiding our prompt design across tasks. The prompt used for this experiment is similar to the \plan One-Shot prompt.

\section{Evaluation Details}
\label{A: Eval}
\definecolor{pback}{HTML}{eeeeee}
\definecolor{pbord}{HTML}{5954a4}
\subsection{Baseline Evaluation Metric: Code Similarity Score}
\label{app:codesim}
A \textbf{Code Similarity} score was also measured for each solution. This is a text-based similarity score between the solution and the GT code. We report this metric as a baseline measure of code similarity, and to further motivate our introduction of domain-specific metrics.

We used the difflib \citep{difflib} package to compute a score between 0 and 1 as a measure of string similarity, using the ratio of the lengths of the longest matched subsequences to the ratio of the lengths of strings being compared. Code Similarity reflects this score between the generated code and the ground truth code. It is not surprising that this metric has the least variation across experiments and models since significantly different blocks of code might yield the same answer.  The preponderance of boilerplate syntax, along with the fact that two different code blocks could generate equivalent model subtrees, are factors that contribute to the lack of meaningful variation of this metric across experiments. As a specific example, a \texttt{model.study("std1").run();} will leverage \comsol's default numerical solver for the problem. However, this could also be represented explicitly using large blocks of \texttt{model.sol("sol1")...} lines in the Ground Truth Code field.

\begin{table}
\centering
\begin{small}
\caption{Code Similarity across LLMs.}
\label{tab:codesim1}
\begin{tabular}{p{1.2in}p{1.8in}p{1in}}
\toprule
Model & \benchmain: \model & \benchlarge \\
\midrule
Claude-3.5-Sonnet & \textbf{0.19}$\pm$0.03 & \textbf{0.20}$\pm$0.01 \\
GPT-4o & 0.17$\pm$0.03 & 0.15$\pm$0.01 \\
Gemini-1.5-Pro & 0.17$\pm$0.03 & 0.15$\pm$0.01 \\
\midrule
Gemma-2-27B-IT & 0.15$\pm$0.02 & \\
Gemma-2-9B-IT & 0.11$\pm$0.02 & \\
CodeGemma-7B-IT & 0.12$\pm$0.02 & \\
\bottomrule
\end{tabular}
\vspace{0.1cm}
\caption{Code Similarity across experiments.}
\label{tab:codesim2}
\begin{tabular}{p{2in}p{1.2in}}
\toprule
Experiment & Code Similarity \\
\midrule
\model: One-Shot & 0.17$\pm$0.03 \\
\model: PhyDoc In-Context & 0.15$\pm$0.02 \\
\model: Multi-Turn Agent & 0.17$\pm$0.03 \\
\midrule
\plan: One-Shot & \textbf{0.21}$\pm$0.03 \\
\plan: PhyDoc In-Context & 0.20$\pm$0.02 \\
\bottomrule
\end{tabular}
\end{small}
\end{table}

\subsection{Executability}
The LLM output is first parsed to identify the block with Java API calls, and further parsed to pythonize the lines (\aref{app:mph}). This filters out lines that are not code or cannot be pythonized and results in a sequence of \comsol API calls and their `pythonized' counterparts, all of which start with \texttt{model.} and end with ';'. 

The pythonized lines are then passed to the MPh client, and replies for each line are received. We parse API replies using the following patterns. A reply containing any of the following [`Messages', `has no attribute', `No matching overloads', `invalid syntax', `Exception', `is not defined'] are considered {\color{red} Syntax Errors}. Replies with [`Ambiguous', `comma', `No Model set'] are {\color{purple} Translation errors}. The last category category is rare in our experiments and are occasionally encountered when we tested adding new problems to the benchmark that contained lines that weren't translated correctly in the query: the first two flag errors in the query to \comsol via Mph, while the last indicates that an action is being done on a non-existent model, which is inconsistent with the setup of the code. All other replies are designated {\color{teal} Correct}.

\begin{equation}
    Executability = \frac{{\color{teal} \rm{Correct Lines}}}{\rm{Total Parsed Lines}}
\end{equation}

\subsection{Model Tree Score}
The model tree representation of the model built by the language model can be extracted, and one can use the same similarity score as above to compute a similarity score relative to the target tree. We expect this to be a more reliable measure of alignment since different blocks of code that build the same model will have the same model tree (addressing the case described in Code Similarity). Using the formula below, the score will be 1.0 if the trees are identical, and 0.0 if the trees are equivalent to a tree before any code is run.

\begin{equation}
    Model Tree Score = \frac{\rm{Score(LM, GT)} - \rm{Score(Empty, GT)} }{1.0 - \rm{Score(Empty, GT)}}
\end{equation}

The following is an empty tree, corresponding to a model that has only been initialized, before any code is run.

\begin{tcolorbox}[width=\textwidth,colback=creamywhite, colframe=black, breakable]
\begin{verbatim}
model
├─ parameters
│  └─ Parameters 1
├─ functions
├─ components
├─ geometries
├─ views
├─ selections
├─ coordinates
├─ variables
├─ couplings
├─ physics
├─ multiphysics
├─ materials
├─ meshes
├─ studies
├─ solutions
├─ batches
├─ datasets
├─ evaluations
├─ tables
├─ plots
└─ exports
\end{verbatim}
\end{tcolorbox}

\subsection{Valid Target}
There are various ways in which computing the correct value and exporting it to a table may fail: a) the LLM's code forgets the export command to the API and no table is exported b) an empty table is exported or, c) a table containing an incorrect value is exported, such as a default value or the wrong quantity (eg: time instead of temperature). Failure modes b) and c) are far more common than a) and occur when the code is not fully correct and the partially constructed \comsol model exports nothing or an incorrect value. For instance, a partially solved model that was asked to compute the temperature at time=190s might export a table where the last value was 190 but because of errors in model construction, no temperature was exported. In such a case if the ground truth answer is say, 185$^\circ$C, without verifying the physical quantity, one would mistakenly evaluate the algorithmically parsed figure 190 to be quite close to the target. In other cases, the software might export a default such as 293.15 K if the solver did not solve correctly. 

If a table containing the target quantity is exported, it is first read and parsed. The last value in the table is algorithmically extracted. To address this problem, we ask an LLM (Gemini-1.5-Pro), to extract the exported value and units from the table, if it is a match for the target description, and minimize the chances of incorrectly evaluating these failure modes as valid solutions. 

\begin{tcolorbox}[width=\textwidth,colback=pback, colframe=pbord, title=Evaluate Prompt, breakable]
You are provided with a table that was exported by a model built in COMSOL. The table * should * contain the EXPECTED TARGET QUANTITY. The following failure modes may occur when the model is not built correctly:

1. The table might be empty or might export a physical quantity that is different from the expected target quantity.

2. The table might export the same physical quantity, but the quantity is just an initial or boundary condition, or a default value that was exported, instead of the result of genuinely numerically solving the problem. You can find numbers already in the problem description in `PROBLEM`. Default values include 20degreesCelsius, 293.15 K, 0 etc.

Carefully examine the `TABLE` and compare it with the units and description of the expected target quantity and the numbers in `PROBLEM` to assess whether the table exported a value that was the result of genuinely numerically solving the problem.
You must return TARGET VALUE and TARGET UNITS in json format if the table was the result of genuinely solving the model, computing a solution and exporting it.
Return `N/A` for both fields if the table suffers from either of the failure modes described above.

\begin{plclisting}
-----
PROBLEM: {{problem_description}}

-----

EXPECTED TARGET QUANTITY: {{target_description}} 

TABLE: {{table}}

REPLY:
\end{plclisting}
\end{tcolorbox}

We then compute the number of problems for which the LM was able to parse the reply and convert it to a JSON. This fraction is the number we report as Valid Target.

\subsection{Relative Error $|$ Strict}
Our strict filter for whether a model has truly solved the problem is to take the subset of problems for which the problem was judged to be a valid export by the LLM, and to consider the algorithmically parsed last value. We then compute the relative error of this value against the ground truth target value. If this value is less than 10\%, we consider it valid.

\subsection{Physics Metrics}
The interface lines are parsed from the ground truth code by finding lines that fit the regex pattern for interface creation. Likewise for the feature creation and feature property modification lines. Each of these lines of codes can be considered as an ``Action" consisting of an Action Type (eg: Create Interface) with corresponding Arguments (eg: Interface tag, Name of the Interface, Geometry).



\textbf{Create Interface:} 
\texttt{model.component("comp1").physics().create("Interface\_tag", "InterfaceName", "Geometry\_tag");}

Eg: \texttt{model.component("comp1").physics().create("ht", "HeatTransfer", "geom1");}

\textbf{Create Feature:} 
\texttt{
model.component("comp1").physics("Interface\_tag").create("\newline Feature\_tag", "FeatureName", Dimension);}

Eg: \texttt{
model.component("comp1").physics("ht").create("temp1", "TemperatureBoundary", 1);}

\textbf{Modify Feature Property:} 
\texttt{model.component("comp1").physics("Interface\_tag")\newline.feature("Feature\_tag").set("Param", "Value");}

Eg: \texttt{model.component("comp1").physics("ht").feature("temp1").set("T0", "1000[degC]");}



\subsubsection{Interface Factuality}
We check whether the Interface name exists in a list of known \comsol interfaces. If it exists in this list, we assign it a factuality of 1, else 0.

\subsubsection{Interface Recall}
How many GT interface creation actions (ignoring Interface\_tag) were also in the LM code? This checks whether the same interface was defined on the same geometry. `nan' if there are no interfaces in the GT (not encountered in our dataset).

\subsubsection{Feature Recall}
Since multiple features may be created under the same interface (eg: 2 Boundary Conditions with different temperatures), we compute the occurrences of \textit{each} GT feature name in the GT code and in the LM code, and a recall for each GT feature name, and then average over all GT features. In our implementation, if no GT features are defined, a) AND no LM features are defined the recall is 1, b) but LM features are defined, the recall is 0.

\subsubsection{Feature Dimension}
Let $F_c$ be all the GT features that are also created by the LM solution. Let $Dim_c$ be the set of $F_c$ such that the LM feature has the same dimension as the GT feature.
Feature Dimension = $\frac{|Dim_c|}{|F_c|}$

This is a correctness and physics reasoning metric as opposed to an alignment-focused metric since creating a TemperatureBoundary with dimension 2 attempts to create a 2D temperature boundary condition. Creating a TemperatureBoundary with dimension 1 attempts to create a temperature on an edge. Thus this measures the LM's ability to correctly deduce the spatial dimension of boundary conditions or other features from the context of the problem.

\subsubsection{Feature Property Recall}
This compares the modify feature property actions. It computes how many GT modify feature property actions were also in the ground truth, \textit{ignoring} differences in Interface\_tag and Feature\_tag. If no GT properties are modified, a) AND no LM features are modified the recall is 1, b) but LM features are modified, the recall is 0.

\section{Querying the \comsol API from Python}
\subsection{The Python-\comsol Bridge}
\label{app:mph}
The raw output of the LLM is a string containing \comsol API commands in Java. An interface between Python and \comsol is needed to execute this code and interact in other ways with the API. We use the Python package MPh \citep{mph} and Rpyc for this. MPh is a scripting interface built on JPype \citep{jp} that enables a Python program to communicate with and build a model in \comsol. Each Java API command in the LM's output can be `pythonized' algorithmically. In most cases, the pythonized line is near identical to the Java line. However, due to differences in Java and Python syntax there exist some corner cases that need to be handled separately. Eg: `new String[]' is exclusively a Java construction, while the notation for booleans in Python is True / False as opposed to true / false in Java. Thus a `pythonizer' is constructed that parses and translates Java API calls to their Python counterparts.

The setup involves the following assumptions: an MPh client object is created. This behaves like a stateful `sandbox', where models can be built by LLMs, code can be evaluated, or information such as the current state of the model tree, properties under a node and the exported table can be queried and retrieved. Although multiple models can be created and set under the client, for simplicity we work with settings that involve a single model. Before running a new solution, the existing model is deleted and a new blank model is created. The LLM actions will modify this blank model. Thus, by design, all lines of code the LLM outputs, should start with `\texttt{model.}' and end with `;'.

\subsection{\comsol Code Structure}
\label{app: codeblocks}
\begin{enumerate}
    \item \textit{Geometry}, if any: This involves identifying the dimensionality of the problem, and constructing a representation of the object being modelled, say a cup, by creating and composing primitive shapes such as ellipses or rectangles to build the object. While already constructed geometries can also be imported from other software such as CAD, in our benchmark, we currently restrict ourselves to models for which we construct the geometry from scratch in \comsol. This typically starts with a `model.component(``comp1").geom' pattern.
    \item \textit{Physics}: This will include specifying all the physical conditions for the problem, including initial or boundary conditions, forces, properties or in the case of mathematics problems, the differential equation. This typically starts with a `model.component(``comp1").physics' pattern. Some problems may additionally have lines that begin with `model.component(``comp`").multiphysics', and set up the coupling between different kinds of physical phenomena. We categorize these lines, if any as `physics' in \fref{fig:blockwise} and \ref{fig:bwise_beforeafter}.
    \item \textit{Material}: Creating materials and assigning them to domains. One can either assign known materials such as `Copper' and the object will inherit the default properties of that material, or define a blank material and its properties such as conductivity from scratch. This typically starts with a `model.component(``comp1").material' pattern.
    \item \textit{Mesh}: Usually a shorter step that involves meshing the surfaces of the geometry to set up elements. This typically starts with a `model.component(``comp1").mesh' pattern.
    \item \textit{Study / Solver:} This involves specifying the conditions of the analysis and solver, such as the number of timesteps. While the solver code can be modified to override defaults, \comsol also has the ability to automatically populate the model with the default solvers most apt for a given problem. This typically starts with a `model.study' or `model.sol' pattern respectively. In \fref{fig:blockwise} and \ref{fig:bwise_beforeafter}, we categorize both patterns as `solver'.
    \item \textit{Results:} Once the numerical solver has completed the analysis, one will likely postprocess the problem, in order to generate desired plots or tables. This typically starts with a `model.result' pattern.
\end{enumerate}

\section{Agent Details}\label{app:agent}
\cready{We design a multi-agent system that interacts with the \comsol API, as well as tools (or specialized functions).} Each agent has a specific role and input context.

\texttt{ControllerAgent}: The main agent that tries to solve the problem description by generating solutions, interacting with the API and calling subagents.\newline
\textbf{Input Context}: Problem description.\newline
\textbf{Components}: Evaluator, ControllerSubAgent\newline
\textbf{Working}: This samples an initial population of \change{N(=20)} solutions using PhyDoc In-Context. Over the course of its trajectory, the agent proposes 40 solutions: 20 from oversampling the initial prompt, and another 20 from correcting the best of the initial 20, and the best solution is selected from the tracked best solutions. This allows us to include gains obtained both from oversampling as well as from correction. For 5 problems, the best solution corresponded to one of the initial population of solutions. Each solution is evaluated by the Evaluator. A fitness score, between 0 and 2, is computed for each solution, using the following formula: Executability + ExportSuccessful where ExportSuccessful is 1 if (the solution computed a value \texttt{AND} had executability above 90\%) and 0 if not. The controller agent tracks a set of best replies using their fitness. The set of best replies stores at least B(=1) solution, as well as all solutions that successfully computed a value. This agent also determines the context to be sent to the CorrectorSubAgent, using the following algorithm:
\begin{itemize}\itemsep0em
    \item Solution to iterate on: We use an iteration criterion inspired by the Markov Chain Monte Carlo (MCMC) acceptance criterion. The solution to iterate on (rendered in the prompt to the CorrectorSubAgent as {\small “CURRENT CODE”}) is (a) the last solution if the last solution has equal fitness as the best solution, and (b) the last solution if a random float between [0, 1] is less than $\alpha = \frac{Last\_Fitness}{Best\_Fitness}$, else the best solution.
    \item ExecutionHistory: The best solutions, if not already used in context upto a maximum of \change{3} best solutions, in addition to the last \change{N\_bad(=1}) replies, if not already in context. \newline
\end{itemize}
\texttt{Evaluator}: This returns the feedback for a solution in a `score' dictionary (Left panel, \fref{fig:tool_demo})\newline
\textbf{Input Context}: An LLM solution. \newline
\textbf{Working}: The evaluator always returns execution feedback and additionally includes subjective feedback from a VerifierLLM if Executability exceeds 90\%. Note, this evaluator is \textit{not} aware of the GT target value.

\texttt{CorrectorSubagent}: This returns an updated solution.\newline
\textbf{Input Context}: Problem description, Current Code and Feedback, Execution History\newline
\textbf{Components}: ToolLookupAgent\newline
\textbf{Working}: This calls the ToolLookupAgent and retrieves its reply. It then includes this reply to the rest of the context received from the ControllerAgent to propose the next solution.

\texttt{ToolLookupAgent:} This calls tools and returns the information retrieved from them. \newline
\textbf{Input Context}: Feedback \newline 
\textbf{Components}: ToolRegistry\newline
\textbf{Working}: The LLM is shown tool descriptions and the input context and must return a list of tool calls, as structured classes using the Langfun \citep{Peng_Langfun_2023} package consisting of the tool name and its arguments. If successfully parsed, each tool is called with its arguments and the replies are concatenated (see \fref{fig:tool_demo} for the feedback and reply for a single step). The tools in the registry are:
\begin{enumerate}\itemsep0em
    \item \texttt{QueryPhysicsInterfaces}: This returns a list of valid physics interfaces.
    \item \texttt{QueryPhysicsFeatures}: This returns the features under an argument \textit{interface} or a list of known features under interfaces.
    \item \texttt{QueryModelTreeProperties}: The LLM must call this tool with a \textit{path} argument (`/physics/Heat Transfer in Solids/Solid 1' in \fref{fig:tool_demo}) to receive the properties under the node corresponding to path.
    \item \texttt{RetrieveAnnotatedSnippets}: To call this tool, the LLM must specify a \textit{branch} -- one of the conceptual blocks such as physics or geometry -- and a \textit{query} -- a brief natural language description of a specific step. In \fref{fig:tool_demo}, the LLM first called this tool with the branch `geometry' and the query `Create a 2D axisymmetric geometry in...'. \change{A retriever then looks up the annotated library and retrieves 3 annotations along with their code snippets, most similar to the query made.} Thus, this allows the LLM to search a library of code snippets to find the correct ways to express certain steps in code, \rewrite{simulating how a human unfamiliar with a coding language would look up similar examples of code.}
\end{enumerate}
At the end of this experiment, the \texttt{ControllerAgent} saves its best solutions as well as other intermediate states. During evaluation, the best solutions are read in and evaluated. If there are multiple best solutions (in cases where multiple solutions were able to compute a target value), the \cready{top} best solution is the one that maximizes the following formula: Executability + bool(Computed Value) + [(1.0 - Target Relative Error) if (Target Relative Error$<$1) AND (Valid Target) else 0]. The three conditions together prioritize solutions that (1) had high executability, (2) were complete enough to export any value, albeit incorrect or the wrong quantity and, (3) exported a `Valid Target' within 10\cready{0}\% of the desired value.

The agent experiment on a single problem takes slightly over 12 minutes (ranging from 7-17 minutes) on average per problem. The dominant factor contributing to this variability is the number of LLM queries: in problems where executability crosses 0.90, there will be more LLM queries since the Evaluator additionally calls the VerifierLLM. The FEA runtime is only a small fraction of this time: parsing the LLM reply, evaluating it by executing it in \comsol and retrieving API messages took around 0.9-1.5s for a single LLM reply. We used a subset of 5 problems to compute these estimates.

\subsection{Tools}
\cready{In our implementation of the ToolLookupAgent, if the tool call fails, the ToolLookupAgent will return an empty reply. Tool calls fail when the LLM is unable to generate a call that is formatted in the way Langfun expects.}
\subsubsection{QueryModelTreeProperties}
In order to help the LLM learn how to appropriately format a valid path, say to the `Solid' feature, the current state of the model tree is shown to the ToolLookupAgent LLM. It also has a history of unsuccessful (incorrectly formatted) paths in previous queries to this tool, in order to minimize the chances of incorrectly calling this tool with an invalid path. 

\subsubsection{RetrieveAnnotatedSnippets}
\label{A:annotated_lib}
We use the Discovery Engine API \citep{retriever} with the model name `semantic-ranker-512-003' to rank and retrieve the top 3 annotations most similar to the query snippet. The annotation library was generated by taking tutorials and splitting them into code blocks using the patterns described in \ref{app: codeblocks}. There are \change{768} pairs of annotations and snippets across all branches of code. Here is an example of an annotation `summary' and its snippet:

\textbf{Summary: }Defining a transient study with a time range from 0 to 0.025 seconds with a step of 1 second. The study will solve for the "spf" physics interface, and a relative tolerance of 0.001 will be used. The number of solver iterations will be automatically determined based on the time step.

\textbf{Code: }
\begin{clisting}
model.study().create("std1");
model.study("std1").create("time", "Transient");
model.study("std1").feature("time").setSolveFor("/physics/spf", true);
model.study("std1").feature("time").set("tlist", "range(0,0.025,1)");
...
model.study("std1").feature("time").set("solnum", "auto");
\end{clisting}

\subsection{Analysis}
\begin{figure}[h]
    \centering
    \includegraphics[width=0.5\textwidth]{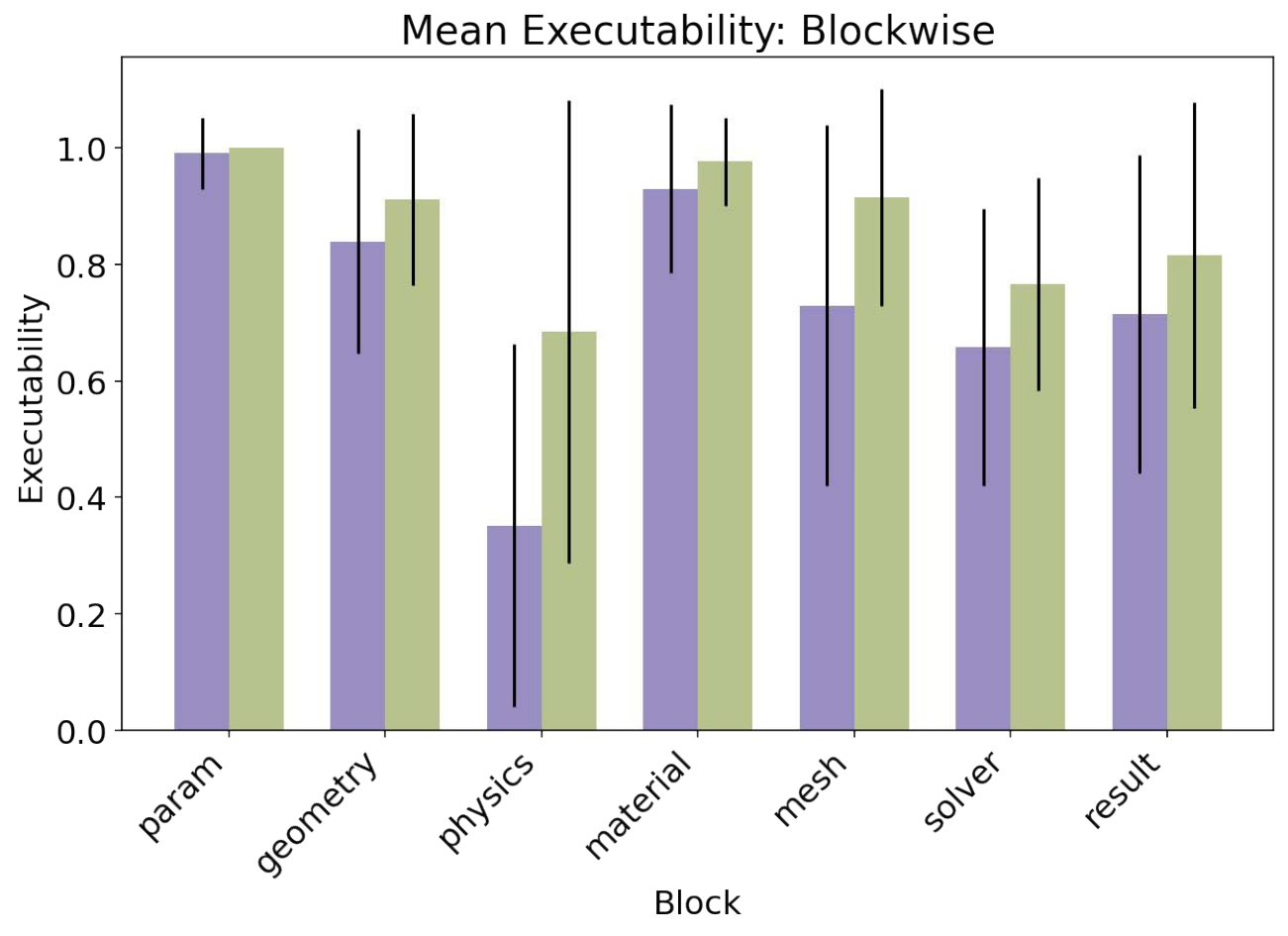}
\caption{Block-wise executability across the 300 initial samples of code (purple) with PhyDoc In-Context and in the best solution (green) across all problems. Error bars denote standard \mbox{deviations}.}
\label{fig:bwise_beforeafter}
\end{figure} 

\begin{figure}[h]
    \centering
    \includegraphics[width=0.4\textwidth]{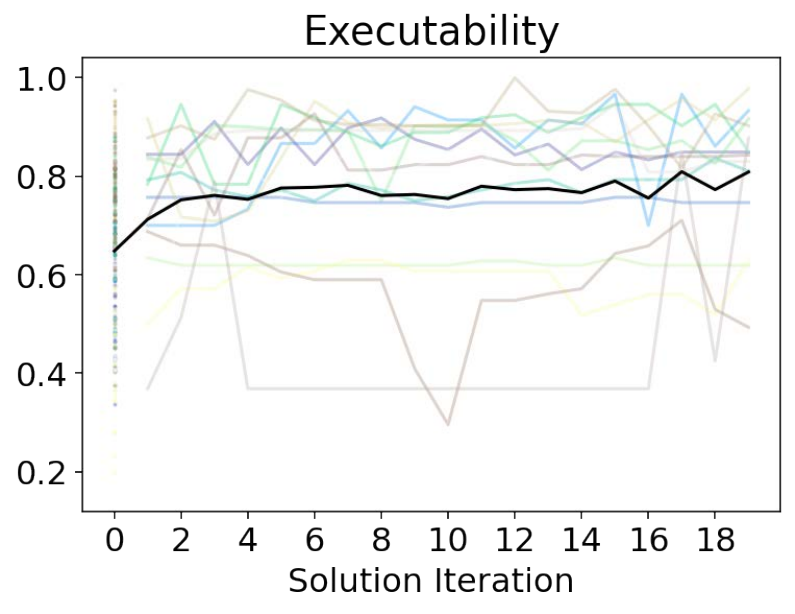}
    \includegraphics[width=0.4\textwidth]{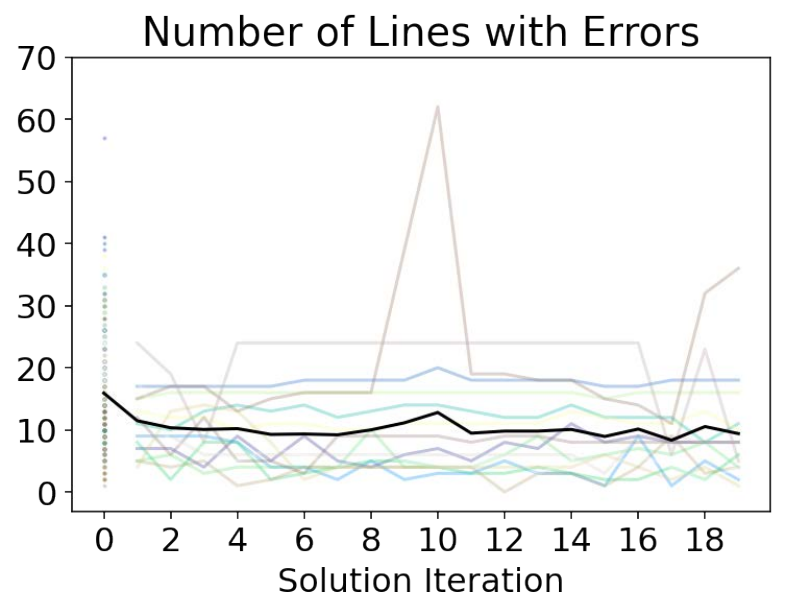}
\caption{Executability and number of errors over solutions returned by the ControllerAgent. The scatter at 0 denotes the spread from the samples in the initial population and the black line denotes the mean value for the metric at that state across all problems. Each colored line demarcates a different problem.}
\label{fig:trajectories}
\end{figure} 

\fref{fig:bwise_beforeafter} depicts the blockwise executability in the initial sample relative to the best solution across problems. The standard deviations in the best case are higher since we have 1 best solution for each problem, and 20 samples per problem in the initial population. \fref{fig:trajectories} plots the Executability as well as the number of errors over solution iteration. The evolution of the metrics isn't monotonic and in some cases the agent gets stuck on the same solution for some iterations, or takes an incorrect turn. We added the acceptance criterion to minimize the number of iterations required to ``escape" an incorrect turn.

\section{Qualitative Analysis}
\label{app:qual}
\begin{figure}[!h]
\includegraphics[width=\textwidth]{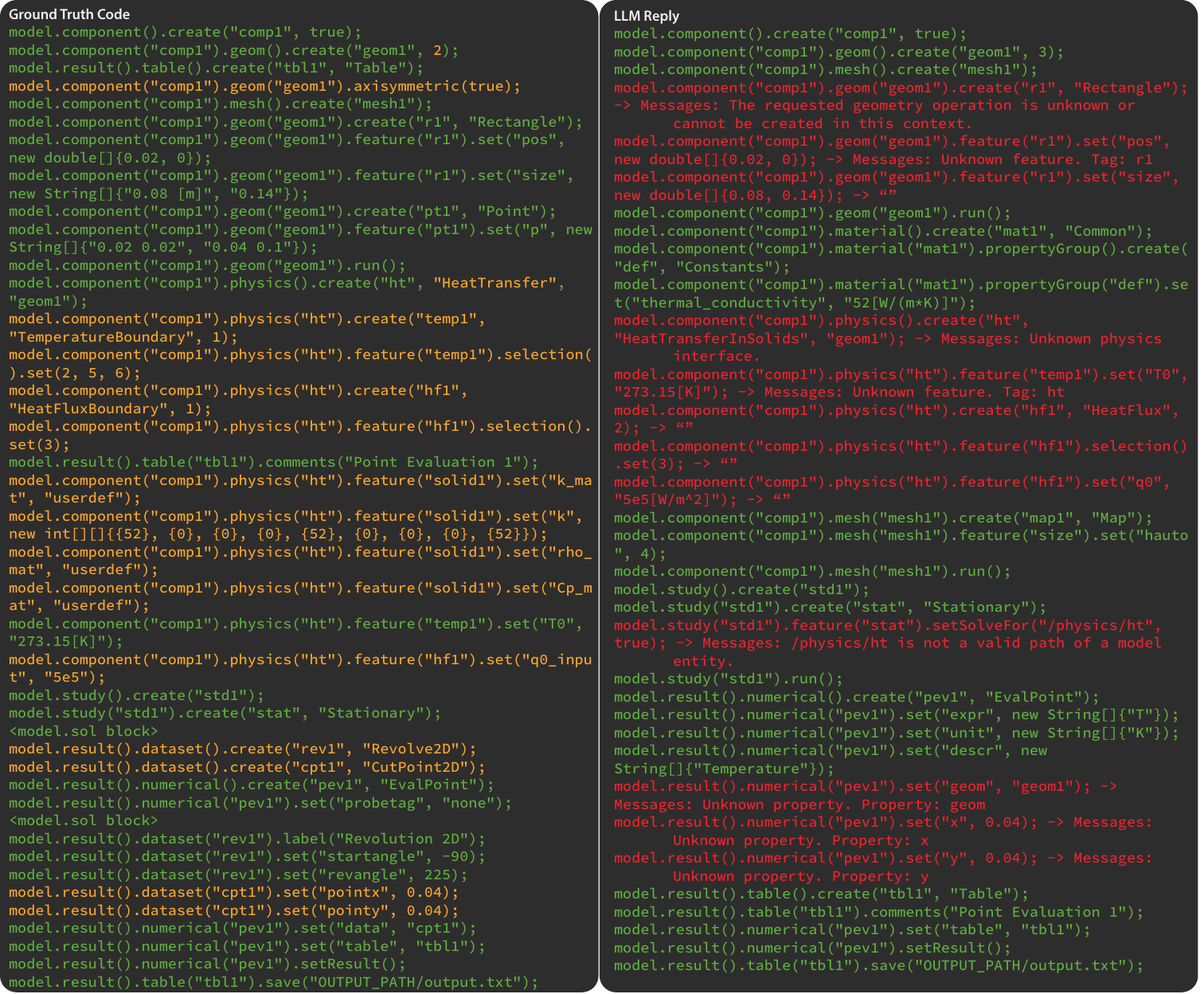}
\caption{Comparing a Ground Truth code with the LLM-generated code. The ochre-colored lines or arguments in the GT code demarcate lines or arguments that were modified or absent in the LLM generated code. The green (red) lines in the LLM-generated code denote lines that were (were not) executable. The arrows against the red lines show the abbreviated API message returned for each non-executable line.}
\label{fig:analysis}
\end{figure}

In \change{\fref{fig:analysis}}, we qualitatively compare the LLM-generated code for the \model task in the baseline (one-shot) setting with Gemini-1.5-Pro, relative to the ground truth code field, for the problem in \ifthenelse{\boolean{owa}}{\fref{fig:schema} and \aref{app: example_bench}
}{
\aref{app: example_bench}}. At a high-level, the LLM's solution consists of API calls that qualitatively possess the same structure and grammar as in the GT code. This problem requires the LLM to represent the cylindrical cross-section of a cylinder as a rectangle in 2D, with the axisymmetric condition applied for rotational symmetry about the cylinder's axis. The LLM instead creates a 3D geometry and attempts to create a rectangle. This doesn't work as is indicated by the error message, since the rectangle is a 2D construct and cannot be directly created as a 3D object. Since the rectangle creation action fails, no `r1' node is created, and subsequent actions acting on the `r1' node are invalid. This pattern of non-executability is also observed downstream, where all actions on the `ht' node are rendered invalid because the `ht' node could not be created in the first place. \verbose{Note, if the LLM had chosen a 2D axisymmetric geometry, the remaining geometry lines of code would be correct. They fail because of an incorrect decision made first.}

The LLM tries to create a `HeatTransferinSolids' interface. We described this pathology in RQ2. In this example, the Interface Recall and Feature Recall metrics are 0, as is the Interface Factuality metric. The GT code modifies 5 features, of which the LLM only modifies 1 (setting $T_0$ to 273.15 K). Thus the Modify Feature Property score is 0.2.

%

\section{Can LLMs solve these problems in Python?}
\label{app:python}
\rewrite{Our experiments underscored the challenge of composing physics reasoning skills with the ability to generate syntactically correct and consistent \comsol API code.} In this section, we seek to evaluate how well LLMs can solve \benchmain { problems} with a general purpose programming language, Python. We set up a Docker environment with the numerical packages, \texttt{numpy} \citep{harris2020array} and \texttt{scipy} \citep{2020SciPy-NMeth}, along with the open-source FEniCSx (\texttt{dolfinx}) \citep{Baratta_DOLFINx_the_next_2023} package. We further test the SWE-Agent \citep{yang2024swe} framework with \texttt{claude-3-5-sonnet-20241022} as the underlying LLM on its ability to solve these problems, since \texttt{claude-3.5-sonnet} generally performed best in \tref{tab:compare_models} and \ref{tab:compare_models2}. 

The problem statement for each problem is in \aref{app:sweagprompts}, with the \textbf{Model Specifications} and the \textbf{Target Description} fields used to substitute the problem\_description and target\_description placeholders respectively. As in the experiments in \comsol, the LLM needs to generate the code that solves the problem, and export the target value and the target units to an `output.txt` file. We then apply the patch containing the code and the output, and evaluate the output. We used a budget limit of \$1 per problem. Note, in this setup, we evaluate the last `output.txt` file. If the agent's attempt at solving the problem terminates because it exceeds the cost limit, the `output.txt` file will correspond to the last code that the agent successfully executed solution in the loop, and \textit{not} the last version of the code.

Although the environment contains FEniCSx, which is likely better suited to solving FEA problems, we explicitly instructed the agent to use only numpy and scipy. We made this choice since preliminary tests revealed that the agent also failed to generate executable FEniCSx code due to breaking changes in the package. 

We evaluate the following metrics:
\begin{itemize}\itemsep0em
    \item \textbf{Valid Target}: As in \sref{sec:eval}, this is the number of problems for which the LLM judged the computed output to be valid i.e. consistent with the target description and units. \textbf{11/15} solutions had a valid target.
   \item \textbf{Relative Error$|$ Strict}: Only \textbf{4} solutions passed the strict cut, i.e. computed a valid target AND a relative error of less than 10\%. The mean relative error over this subset is $\mathbf{2.15\% \pm 1.89\%}$. 
\end{itemize}
Over the larger set of 11 problems for which any solution was computed, the mean relative error is $296\%\pm200\%$, where the error bars denote the standard error on the mean.

In Python, and particularly with popular `in-distribution' packages like numpy and scipy, the difficulty of generating syntactically correct code is minimal. However, now the LLM must define and implement the equations describing the physics and the mesh from scratch and configure its own solver settings as it can no longer import pre-verified physics modules and automatic solvers as in the engineering simulation software. \rewrite{Thus the bottleneck has now shifted from generating executable code, to being able to achieve the desired numerical precision, since in only 4 cases the LLM was able to compute a value within 10\% of the desired target.} The emergence of commercial softwares like \comsol, Ansys\regc \citep{ansys} and Abaqus\regc FEA \citep{abaqus} arose from the challenge of solving complex engineering problems with general purpose languages, and to fulfil the need for already validated physics modules, numerical precision and automatic solvers.

\section{Prompts}
\label{app: prompts}
\subsection{Single Query Prompts}
\definecolor{pback}{HTML}{eeeeee}
\definecolor{pbord}{HTML}{5954a4}

\begin{tcolorbox}[width=\textwidth,colback=pback, colframe=pbord, title=\model $|$ One-Shot, breakable]
You are an experienced COMSOL engineer. You must solve the problem to compute the desired TARGET QUANTITY by generating COMSOL JAVA API code.
The model creation line ```Model model = ModelUtil.create("Model");''' has already been generated and you should not repeat this line.
All lines of code must begin with `model.`

You must not generate any `model.sol...` solver code but should ensure that your `model.study...` block ends with a `model.study("std1").run();`. This will automatically create and run the default solver for the problem.
Use the example provided below to infer how to format your response and generate COMSOL code.
===

EXAMPLE 0:

\textbf{PROBLEM DESCRIPTION:}
\#\# Stress Analysis of an Elliptic Membrane

**ANALYSIS TYPE:**

* Linear elastic, Plane Stress.

**GEOMETRY:**

* The domain is a quarter of an elliptical membrane.

* The outer curved edge is defined by the equation: $(x/3.25)^2 + (y/2.75)^2 = 1$

* The inner curved edge is defined by the equation: $(x/2)^2 + y^2 = 1$

* Thickness: 0.1 meters (uniform throughout)

* Labeled points:
    * Bottom Left Corner, Point O: (x = 2.0, y= 0)

**LOADING:**

* Uniform outward pressure of 10 MPa is applied on the outer curved edge, normal to the boundary.
* The inner curved edge is unloaded.

**BOUNDARY CONDITIONS:**

* Left Edge: Symmetry about the y-axis, implying zero displacement in the x-direction.
* Bottom Edge: Symmetry about the x-axis, implying zero displacement in the y-direction.

**MATERIAL PROPERTIES:**

* Isotropic: The material properties are the same in all directions.
* Young's Modulus (E): 2.1 x 10$^11$ Pa
* Poisson's Ratio ($\nu$): 0.3

**ELEMENT TYPES:**

* Plane stress: The analysis assumes the membrane is thin and subjected to in-plane loading.

**MESHES:**

* A mapped quadrilateral mesh over the entire Quarter-Symmetry Domain.

**OUTPUT:**

* The analysis aims to calculate the tangential edge stress ($\sigma_{yy}$) at point O.

Export the table containing this value to OUTPUT\_PATH/output.txt.

\textbf{SELECTION IDENTITIES:}

DOMAINS:
Quarter-symmetry membrane: Domain 1

BOUNDARIES:
Left edge, Symmetry Boundary Condition: Boundary 1

Bottom edge, Symmetry Boundary Condition: Boundary 2

Outer Curved Edge, Boundary Load: Boundary 4

POINTS:

O: Point 3

TARGET QUANTITY to be computed: Tangential stress ($\sigma_{yy}$) at point O in Pa.
Export the table containing this value to OUTPUT\_PATH/output.txt.

\textbf{ANSWER:}
\begin{clisting}
java```
model.component().create("comp1", true);
model.component("comp1").geom().create("geom1", 2);
model.component("comp1").mesh().create("mesh1");

model.component("comp1").geom("geom1").create("e1", "Ellipse");
model.component("comp1").geom("geom1").feature("e1").set("semiaxes", new double[]{3.25, 2.75});
model.component("comp1").geom("geom1").feature("e1").set("angle", 90);
model.component("comp1").geom("geom1").create("e2", "Ellipse");
model.component("comp1").geom("geom1").feature("e2").set("semiaxes", new int[]{2, 1});
model.component("comp1").geom("geom1").feature("e2").set("angle", 90);
model.component("comp1").geom("geom1").create("dif1", "Difference");
model.component("comp1").geom("geom1").feature("dif1").selection("input").set("e1");
model.component("comp1").geom("geom1").feature("dif1").selection("input2").set("e2");
model.component("comp1").geom("geom1").run();
model.component("comp1").geom("geom1").run("fin");

model.component("comp1").selection().create("sel1", "Explicit");
model.component("comp1").selection("sel1").geom("geom1", 0);
model.component("comp1").selection("sel1").set(3);
model.component("comp1").selection("sel1").label("O");

model.component("comp1").material().create("mat1", "Common");
model.component("comp1").material("mat1").propertyGroup().create("Enu", "Young's modulus and Poisson's ratio");
model.component("comp1").material("mat1").propertyGroup("Enu").set("E", "2.1e11");
model.component("comp1").material("mat1").propertyGroup("Enu").set("nu", "0.3");

model.component("comp1").physics().create("solid", "SolidMechanics", "geom1");
model.component("comp1").physics("solid").create("sym1", "SymmetrySolid", 1);
model.component("comp1").physics("solid").feature("sym1").selection().set(2);
model.component("comp1").physics("solid").create("sym2", "SymmetrySolid", 1);
model.component("comp1").physics("solid").feature("sym2").selection().set(1);
model.component("comp1").physics("solid").create("bndl1", "BoundaryLoad", 1);
model.component("comp1").physics("solid").feature("bndl1").selection().set(4);


model.component("comp1").physics("solid").prop("Type2D").set("Type2D", "PlaneStress");
model.component("comp1").physics("solid").prop("d").set("d", ".1");
model.component("comp1").physics("solid").feature("bndl1").set("LoadType", "FollowerPressure");
model.component("comp1").physics("solid").feature("bndl1").set("FollowerPressure", "-10 [MPa]");

model.component("comp1").mesh("mesh1").create("map1", "Map");
model.component("comp1").mesh("mesh1").feature("size").set("hauto", 4);
model.component("comp1").mesh("mesh1").run();

model.study().create("std1");
model.study("std1").create("stat", "Stationary");
model.study("std1").feature("stat").setSolveFor("/physics/solid", true);
model.study("std1").run();

model.result().numerical().create("pev1", "EvalPoint");
model.result().numerical("pev1").selection().named("sel1");
model.result().numerical("pev1").set("probetag", "none");

model.result().table().create("tbl1", "Table");
model.result().table("tbl1").comments("Point Evaluation 1");
model.result().numerical("pev1").set("table", "tbl1");
model.result().numerical("pev1").set("expr", new String[]{"solid.syy"});
model.result().numerical("pev1").set("unit", new String[]{"N/m^2"});
model.result().numerical("pev1").set("descr", new String[]{"Stress tensor, yy-component"});
model.result().numerical("pev1")
      .set("const", new String[][]{{"solid.refpntx", "0", "Reference point for moment computation, x-coordinate"}, {"solid.refpnty", "0", "Reference point for moment computation, y-coordinate"}, {"solid.refpntz", "0", "Reference point for moment computation, z-coordinate"}});*
model.result().numerical("pev1").setResult();
model.result().table("tbl1").save("OUTPUT\_PATH/output.txt");
```
\end{clisting}
===

Now generate the JAVA API code to compute the target quantity for the problem below. Export the table containing the target quantity to OUTPUT\_PATH/output.txt.
\begin{plclisting}
PROBLEM DESCRIPTION: {{problem_description}}

TARGET QUANTITY to be computed: {{target_description}}

ANSWER:
\end{plclisting}
\end{tcolorbox}

We used the prompt above for the \model experiment with Gemini-1.5-Pro. 
We used an identical prompt for Claude-3.5-Sonnet and GPT-4o with the 3rd last line of code in the One-Shot example (marked by *) deleted, since it prevented us from querying those LLMs. We also used the version of the prompt with this line deleted for the Agent experiment using Gemini-1.5-Pro.

\begin{tcolorbox}[width=\textwidth,colback=pback, colframe=pbord, title=\plan $|$ One-Shot, breakable]
You are an experienced COMSOL engineer. You must generate the COMSOL API code in JAVA to execute the steps described in the plan below to compute the desired TARGET QUANTITY by generating COMSOL JAVA API code.
The model creation line ```Model model = ModelUtil.create("Model");''' has already been generated and you should not repeat this line.
All lines of code must begin with `model.`
You must not generate any `model.sol...` solver code but should ensure that your `model.study...` block ends with a `model.study("std1").run();`. This will automatically create and run the default solver for the problem.

Use the example provided below to infer how to format your response and generate COMSOL code.

===

EXAMPLE 0:

\textbf{PLAN:} \#\# Implementing the Elliptic Membrane Analysis in COMSOL Multiphysics:

**1. Model Setup:**

* **New Model:** Start COMSOL Multiphysics and create a new model.

* **Space Dimension:** Select 2D for the space dimension.

* **Physics Selection:** Choose the "Structural Mechanics Module" and select "Solid Mechanics" as the physics interface.

* **Study:** Create a new "Stationary" study.

**2. Geometry Creation:**

* **Geometry Primitives:** Use the "Ellipse" tool to create two quarter ellipses representing the outer and inner boundaries. To get a quarter-symmetry geometry, limit the sector angle to 90 degrees.

    * Outer Ellipse: Center (0, 0), Semi-axes (3.25, 2.75) meters, sector angle = 90 degrees.
    
    * Inner Ellipse: Center (0, 0), Semi-axes (2, 1) meters, sector angle = 90 degrees.
    
* **Boolean Operations:** Use the "Difference" operation to subtract the inner ellipse from the outer ellipse, creating the quarter-symmetry membrane geometry.

**3. Definitions:**

* **Points:** Create an explicit selection for Point O (Point 3).

**4. Material Properties:**

* **Material Definition:** In the "Material" node, define a new material with the following properties:

    * Young's Modulus (E): 2.1e11 Pa
    
    * Poisson's Ratio ($\nu$): 0.3

**5. Physics:**

* **2D Approximation:** Use the "Plane Stress" physics approximation, with a thickness of 0.1 meters.

**6. Boundary Conditions:**

* **Symmetry:**
    * Select the bottom edge (Boundary 2) and apply a "Symmetry" boundary condition.
    
    * Repeat the same for the left edge (Boundary 1).
    
* **Pressure Load:** Pressure load of 10e6 Pa acting outwards.
    * Select the outer curved edge Boundary 4 and apply a "Boundary Load" boundary condition with a "Pressure load" of magnitude of -10 MPa.

**7. Meshing:**
* **Mesh Creation:** Right-click on the "Mesh" node and choose "Mapped".
* **Mesh Size:** Adjust the mesh size settings to "Fine".

**8. Study Setup:**
* **Study Type:** Choose a "Stationary" study to analyze the static equilibrium state.
* **Solver Configuration:** Use the default solver settings.

**9. Solving the Model:**
* **Compute:** Click on the "Compute" button to run the finite element analysis.

**10. Post-Processing:**
* **Point Evaluation:**
    * Add a "Point Evaluation" node to extract the tangential stress ($\sigma_{yy}$) at point O.
    * Select point O.
    * Evaluate the expression "solid.syy".
    * Export the table containing this value to OUTPUT\_PATH/output.txt.

\textbf{TARGET QUANTITY to be computed:} Tangential edge stress $\sigma_{yy}$) at O in Pa.

\textbf{ANSWER:} 
\begin{clisting}
java```
<<SAME AS CODE IN MODELSPECS ONE-SHOT PROMPT>>
```
\end{clisting}
===

Now generate the JAVA API code to compute the target quantity for the problem below, by following the plan described. Export the table containing the target quantity to OUTPUT\_PATH/output.txt.
\begin{plclisting}
PLAN: {{problem_description}}
TARGET QUANTITY to be computed: {{target_description}}
ANSWER:
\end{plclisting}
\end{tcolorbox}
We used the prompt above for the \plan experiment on Gemini-1.5-Pro

\begin{tcolorbox}[width=\textwidth,colback=pback, colframe=pbord, title=\model $|$ Phy-Doc In-Context, breakable]
You are an experienced COMSOL engineer. You must solve the problem to compute the desired TARGET QUANTITY by generating COMSOL JAVA API code.
The model creation line ```Model model = ModelUtil.create("Model");``` has already been generated and you should not repeat this line.
All lines of code must begin with `model.`
You must not generate any `model.sol...` solver code but should ensure that your `model.study...` block ends with a `model.study("std1").run();`. This will automatically create and run the default solver for the problem.

You are provided with the list of valid physics interfaces and valid features under interfaces. You must only use the interfaces in the available interfaces list.

===

AVAILABLE COMSOL PHYSICS INTERFACES: 
\begin{clisting}
['BeamCrossSection', 'PorousMediaFlowRichards', 'MoistureTransportInBuildingMaterials', 'CreepingFlow', 'CathodicProtection'... <List of 140 Interface>...'LumpedBattery', 'CompressiblePotentialFlow', 'BatteryBinaryElectrolyte', 'ColdPlasma', 'LaplaceEquation', 'DilutedSpeciesInPorousCatalysts']
\end{clisting} 

AVAILABLE FEATURES UNDER INTERFACES:
\begin{clisting}
{'ElectromagneticWavesBeamEnvelopes': {'features': ['MatchedBoundaryCondition', 'SymmetryPlane', 'Scattering', 'TransitionBoundaryCondition', 'Impedance', 'Port', 'FieldContinuity'], 'physics_tags': ['ewbe']}, 'TransientPressureAcoustics': {'features': ['InteriorSoundHard', 'InteriorLumpedSpeakerBoundary', 'TransientMonopoleLineSource', 'CylindricalWaveRadiation', 'Impedance', 'NonlinearAcousticsWestervelt', 'Pressure', 'PlaneWaveRadiation'], 'physics_tags': ['actd', 'actd2']}, ...<Interface-Feature Mapping>...'PressureAcousticsAsymptoticScattering': {'features': [], 'physics_tags': ['paas']}, 'ElectromagneticWavesBoundaryElements': {'features': [], 'physics_tags': ['embe']}, 'WallDistance': {'features': ['Wall'], 'physics_tags': ['wd', 'wd2']}}
\end{clisting}

===

Use the example provided below to infer how to format your response and generate COMSOL code.

===

EXAMPLE 0: $<$Same Example as in the \model One-Shot Prompt$>$

===
Now generate the JAVA API code to compute the target quantity for the problem below. Export the table containing the target quantity to OUTPUT\_PATH/output.txt.

\begin{plclisting}
PROBLEM DESCRIPTION: {{problem_description}}
TARGET QUANTITY to be computed: {{target_description}}
ANSWER:
\end{plclisting}

\end{tcolorbox}
We use the prompt above for the \model + PhyDoc experiment, as well as to sample the initial population in the Multi-Turn Agent experiment. In the latter case, we removed the 3rd last line of code in the One-Shot example.

\begin{tcolorbox}[width=\textwidth,colback=pback, colframe=pbord, title=\plan $|$ Phy-Doc In-Context, breakable]
You are an experienced COMSOL engineer. You must generate the COMSOL API code in JAVA to execute the steps described in the plan below to compute the desired TARGET QUANTITY by generating COMSOL JAVA API code.
The model creation line ```Model model = ModelUtil.create("Model");``` has already been generated and you should not repeat this line.
All lines of code must begin with `model.`
You must not generate any `model.sol...` solver code but should ensure that your `model.study...` block ends with a `model.study("std1").run();`. This will automatically create and run the default solver for the problem.

You are provided with the list of valid physics interfaces and features under each interface. You must only use the interfaces and features in these lists:

===

AVAILABLE COMSOL PHYSICS INTERFACES:
\begin{clisting}
['BeamCrossSection', 'PorousMediaFlowRichards', 'MoistureTransportInBuildingMaterials', 'CreepingFlow', 'CathodicProtection'... <List of 140 Interface>...'LumpedBattery', 'CompressiblePotentialFlow', 'BatteryBinaryElectrolyte', 'ColdPlasma', 'LaplaceEquation', 'DilutedSpeciesInPorousCatalysts']
\end{clisting} 

AVAILABLE FEATURES UNDER EACH INTERFACE: 
\begin{clisting}
{'ElectromagneticWavesBeamEnvelopes': {'features': ['MatchedBoundaryCondition', 'SymmetryPlane', 'Scattering', 'TransitionBoundaryCondition', 'Impedance', 'Port', 'FieldContinuity'], 'physics_tags': ['ewbe']}, 'TransientPressureAcoustics': {'features': ['InteriorSoundHard', 'InteriorLumpedSpeakerBoundary', 'TransientMonopoleLineSource', 'CylindricalWaveRadiation', 'Impedance', 'NonlinearAcousticsWestervelt', 'Pressure', 'PlaneWaveRadiation'], 'physics_tags': ['actd', 'actd2']}, ...<Interface-Feature Mapping>...'PressureAcousticsAsymptoticScattering': {'features': [], 'physics_tags': ['paas']}, 'ElectromagneticWavesBoundaryElements': {'features': [], 'physics_tags': ['embe']}, 'WallDistance': {'features': ['Wall'], 'physics_tags': ['wd', 'wd2']}}
\end{clisting}

===

Now use the example provided below to infer how to format your response and generate COMSOL code.

===

EXAMPLE 0:
PLAN:
...$<$Same as the One-Shot Example in \plan above$>$...

===

===
Now generate the JAVA API code to compute the target quantity for the problem below, by following the plan described. Export the table containing the target quantity to OUTPUT\_PATH/output.txt.
\begin{plclisting}
PLAN: {{problem_description}}
TARGET QUANTITY to be computed: {{target_description}}
ANSWER:
\end{plclisting}
\end{tcolorbox}

\subsection{Multi-Turn Agent Prompts}
The following prompt is used in the ToolLookupAgent to call tools. \texttt{tool\_snippet} is populated with the descriptions of each tool. \texttt{state\_info} is the execution and verifier feedback for the solution to iterate upon (left panel of \fref{fig:tool_demo}).
\begin{tcolorbox}[width=\textwidth,colback=pback, colframe=pbord, title=Tool Selection, breakable]
You are a COMSOL engineer. You are attempting to gather information relevant to execution feedback that you received from the COMSOL client after you executed some code. The relevant information can be queried as `ToolCall`. Each `ToolCall` must consist of str along with the relevant arguments, if any. A ToolCall may or may not require arguments. Identify the relevant tool calls and return your reply as a `ToolCalls` object, which consists of a list of `ToolCall`s.

===

Here is some information on each tool
\begin{plclisting}
{{tool_snippet}}
\end{plclisting}

===

Now return the relevant ToolCallList for the following execution feedback / error message.
\begin{plclisting}
FEEDBACK: {{state_info}}
\end{plclisting}
\end{tcolorbox}

\lstnewenvironment{txlisting}{\lstset{basicstyle=\rmfamily\small,
breaklines=true}}{}

\begin{tcolorbox}[width=\textwidth,colback=pback, colframe=pbord, title=Correction Prompt, breakable]
You are an engineer solving the following PROBLEM in COMSOL, by generating a solution that consists of the JAVA COMSOL API code needed to solve the problem. You have so far generated the code in CODE.
On executing the lines in CODE you encountered the issue described in CURRENT EXECUTION FEEDBACK.
CURRENT EXECUTION FEEDBACK is formatted as `Line $\rightarrow$ Status: Error (if Status=`Error`)` where Status is `Correct` if the line of code was able to execute and `Error` if it raised an error.
You have additionally been provided with EXECUTION HISTORY which is a record of some of your previous code solutions and their execution results. You may use it as relevant context to understand what blocks of code work and what you've already tried.

You must return a BETTER solution by correcting lines of code that raised errors, or substituting blocks of code with other equivalent code snippets that would solve the problem. The solution must be a full contiguous block of CODE.
Use the example provided below to understand how to format your CODE.

===

EXAMPLE 0:

PROBLEM:* Select 2D for the space dimension.

* Select Fluid Flow $>$ Single-Phase Flow $>$ Laminar Flow (spf).

* Create a Stationary Study

* Insert a geometry from file.

**Parameters**

\begin{lstlisting}[basicstyle=\ttfamily, breaklines=true]
* Name & Expression & Description \\
* Re & 100 & Reynolds number \\
* rho0 & 1e3 [kg/m$^3$] & Density \\
\end{lstlisting}

CODE:
\begin{clisting}
java```
model.component().create("comp1", true);

model.component("comp1").geom().create("geom1", 2);

model.component("comp1").mesh().create("mesh1");

model.component("comp1").physics().create("spf", "FluidFlow", "geom1");

model.study().create("std1");
model.study("std1").create("stat", "Stationary");
model.study("std1").feature("stat").setSolveFor("/physics/spf", true);
model.study("std1").run();
model.component("comp1").geom("geom1").insertFile("fname.mph", "geom1");
model.component("comp1").geom("geom1").run("fin");

model.param().label("Geometrical Parameters");
model.param().create("par2");
model.param("par2").set("Re", "100");
model.param("par2").descr("Re", "Reynolds number");
model.param("par2").set("rho0", "1e3[kg/m^3]");
model.param("par2").descr("rho0", "Density");
...
```
\end{clisting}

EXECUTION HISTORY: 
CURRENT EXECUTION FEEDBACK:
\begin{clisting}
model.component().create("comp1", true); -> Correct
model.component("comp1").geom().create("geom1", 2); -> Correct
model.component("comp1").mesh().create("mesh1"); -> Correct
model.component("comp1").physics().create("spf", "FluidFlow", "geom1"); -> Error: Exception com.comsol.util.exceptions.FlException: Unknown Interface
Messages:
	Unknown Interface
	- Interface: FluidFlow
...
\end{clisting}
The following information may be useful to you:

RELEVANT INFORMATION: -

NEW CODE: 
The Interface `FluidFlow` is not a valid physics interface. LaminarFlow is a valid COMSOL physics interface. 
I will replace FluidFlow with LaminarFlow and return the entire code block.
\begin{clisting}
java```
model.component().create("comp1", true);
model.component("comp1").geom().create("geom1", 2);
model.component("comp1").mesh().create("mesh1");
model.component("comp1").physics().create("spf", "LaminarFlow", "geom1");
model.study().create("std1");
model.study("std1").create("stat", "Stationary");
model.study("std1").feature("stat").setSolveFor("/physics/spf", true);
model.study("std1").run();
model.component("comp1").geom("geom1").insertFile("fname.mph", "geom1");
model.component("comp1").geom("geom1").run("fin");

model.param().label("Geometrical Parameters");
model.param().create("par2");
model.param("par2").set("Re", "100");
model.param("par2").descr("Re", "Reynolds number");
model.param("par2").set("rho0", "1e3[kg/m^3]");
model.param("par2").descr("rho0", "Density");
```
\end{clisting}
===

Here are some example errors, their causes, and example actions that should be taken to address them:

1. Error: `Unknown feature`...
Cause: The feature either does not exist, or is created under the wrong node. It's possible that a feature may be a defined under another feature of the interface, instead of under the interface directly. Eg: `model.component("comp1").physics("int1").feature("f2")...` might raise an error because the correct pattern is `model.component("comp1").physics("int1").feature("f1").feature("f2")...`
Action: Ensure the feature actually exists and substitute it with a similar sounding feature if it doesn't, or define it under the correct node.

2. Error: `Undefined material property 'A' required by FeatureNode F. 
Cause: An essential property needed by F (usually a solver/physics node) has not been defined correctly.
Action: Edit the code where `A` is defined. Try to set the property in one of the following ways instead.
a) Easier Way. You can define a "userdefined" property under the appropriate feature branch of the `physics` branch. The code in this case looks like:

\begin{clisting}
```
model.component("comp1").physics("int1").feature("f1").set("A", "userdef");
model.component("comp1").physics("int1").feature("f1").set("A", "A_value");
```
\end{clisting}

You must have the first line, that sets the property to `userdef` in this case, otherwise f1 might not be able to see A\_value.

b) Harder Way. The property value is defined under the appropriate propertygroup of the material. The code should look like this:

\begin{clisting}
```model.component("comp1").material("mat1").propertyGroup("def").set("density", "7200");```
\end{clisting}

If the property is defined under another propertygroup of the material, the physics branch will sometimes not know where to look, and the code could fail silently.

3. Error: The code saves a value but it's far from the expected value, even though the code is executable. 
Cause: There might be an issue with the study code. You might be missing study settings or the `study.run();` line which is essential for the default numerical solver to run. You should also preferably not generate any `model.sol` lines and ensure that the `model.study..` block ends with `model.study.run();` as this automatically chooses the default COMSOL solver for the problem and runs it.
Action: Try to redefine the .study() code so it includes only the bare minimum described in `Cause`.

4. Error: `Feature cannot be created in dimension`. 
Cause: The feature is being created in a dimension inconsistent with the dimension of the problem. 
Action: Examine what the dimension of the goemetry is and reassess what the correct dimension of the feature should be. For example, a domain feature will typically have the same dimension as the geometry and a boundary feature will have D\_geom -1.  

5. Error: `SelectionOutOfBoundsException: Illegal input vector illegal entity number.` 
Cause: An incorrect or non-existent entity number has been assigned. 
Action: Please recheck the SELECTION INFORMATION and ensure your code is exactly consistent with it.

Note, this is NOT an exhaustive list, and several other errors can occur. Read the error messages carefully, as they typically provide hints about the cause.

===

Now return the corrected code for the following problem:
\begin{plclisting}
PROBLEM: {{problem}}

EXECUTION HISTORY: {{history}}

CURRENT CODE:
```
{{code}}
```

CURRENT EXECUTION FEEDBACK: {{state_info}}
\end{plclisting}
The following information may be useful to you:

\begin{plclisting}
RELEVANT INFORMATION: {{tool_lookup}}

CORRECTED CODE:
\end{plclisting}
\end{tcolorbox}

\subsection{SWE-Agent Experiment Prompts}
\label{app:sweagprompts}
\begin{tcolorbox}[width=\textwidth,colback=pback, colframe=pbord, title=Problem Statement, breakable]
You are provided with a PROBLEM DESCRIPTION and a TARGET DESCRIPTION. This is a problem that will likely require solving the partial differential equations that describe the system using finite element analysis. You may use numpy or scipy, which have been preinstalled. Do not use any other packages unless absolutely necessary. First generate the Python code to solve this problem in `main.py`. 

You must then execute the code and compute the target value corresponding to TARGET DESCRIPTION. This target value needs to be exported to a file called `output.txt` along with its units. For example, if the target value you compute is `30 K`, `output.txt` should have a single line saying `30 K`. If the code executed successfully, read in the `output.txt` file and check whether the computed value seems reasonable according to your analytical guess.
\begin{plclisting}
PROBLEM DESCRIPTION: {{problem_description}}
TARGET DESCRIPTION: {{target_description}}
\end{plclisting}
\end{tcolorbox}

\begin{tcolorbox}[width=\textwidth,colback=pback, colframe=pbord, title=Evaluate Target Validity, breakable]
You are provided with the contents of a file exported by Python code used to solve a problem. The contents * should * contain the EXPECTED TARGET QUANTITY. If the code is incorrect, the contents of the file may be empty or might export a physical quantity that is different from the expected target quantity.

Carefully examine the `OUTPUT` and compare it with the units and description of the expected target quantity and the numbers in `PROBLEM` to assess whether the table exported a value that was the result of genuinely numerically solving the problem.
You must return TARGET VALUE and TARGET UNITS in json format if the output was the result of genuinely solving the problem, computing a solution and exporting it.
Return `N/A` for both fields, otherwise.

-----
\begin{plclisting}
PROBLEM: {{problem_description}}

-----

EXPECTED TARGET QUANTITY: {{target_description}} 

OUTPUT: {{output}}

REPLY:
\end{plclisting}
\end{tcolorbox}

\end{document}